%% file: main.tex
\documentclass[runningheads]{llncs}

\usepackage{eccv}

\usepackage{eccvabbrv}

\usepackage{graphicx}
\usepackage{booktabs}

\usepackage[accsupp]{axessibility}  

\usepackage[width=122mm,left=12mm,paperwidth=146mm,height=193mm,top=12mm,paperheight=217mm]{geometry}
\usepackage{hyperref}

\usepackage{orcidlink}

\usepackage{arydshln}
\usepackage{algorithm}
\usepackage{algpseudocode}
\usepackage{multirow}
\usepackage{makecell}

\usepackage{booktabs} 
\usepackage{array}    

\newcommand{\method}{EMCID}
\makeatletter
\def\blfootnote{\xdef\@thefnmark{}\@footnotetext}
\makeatother

\begin{document}

\title{Editing Massive Concepts in Text-to-Image Diffusion Models} 


\author{Tianwei Xiong\inst{1,2*} \and
Yue Wu\inst{3^*} \and
Enze Xie\inst{4\dagger} \and \\
Yue Wu\inst{4} \and
Zhenguo Li\inst{4} \and
Xihui Liu\inst{1\dagger}
}

\authorrunning{T.~Xiong et al.}

\institute{$^{1}$The University of Hong Kong ~~~ $^{2}$Tsinghua University ~~~ $^{3}$Peking University
~~~ $^{4}$Huawei Noah's Ark Lab
}


\maketitle
\blfootnote{*Equal contribution.~~$\dagger$Corresponding authors.}
\input{sec/0_abstract}
\begin{figure}[]
  \centering
  \vspace{-0.18in}
  \includegraphics[width=\linewidth]{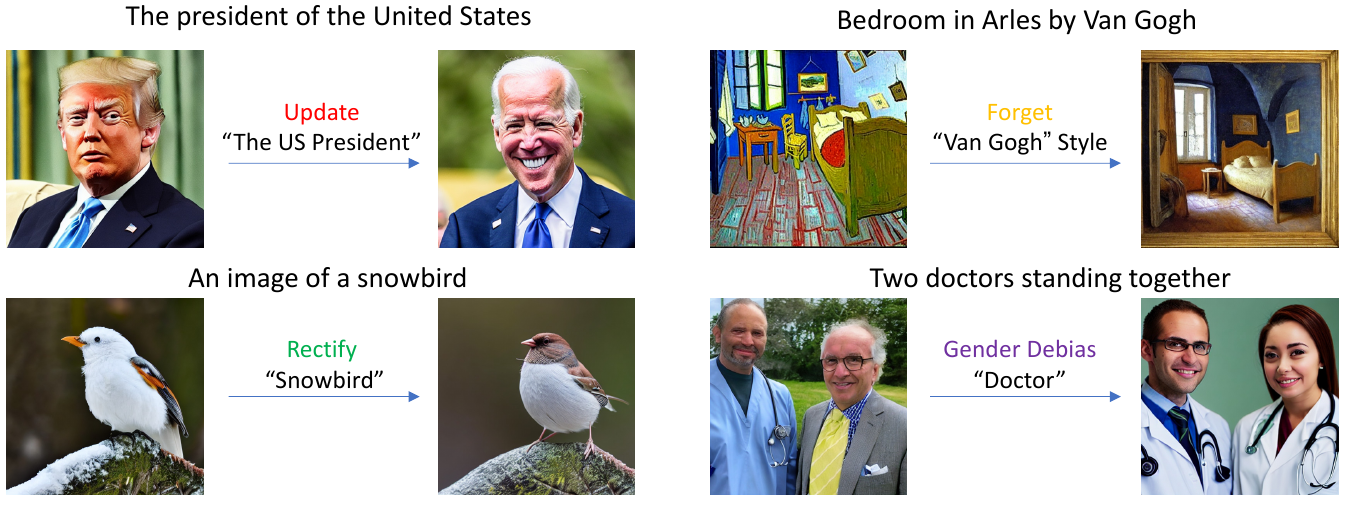}
  \caption{Our method \method{} generally edits \textbf{source concepts}, the concepts intended to be modified, to match \textbf{destination concepts}, the concepts towards which source concepts are to be altered. Our method can update, forget, rectify, and debias various concepts simultaneously at a large scale. 
  }
  \vspace{-0.20in}
  \label{fig:mainfig}
\end{figure}
\input{sec/1_intro}

\input{sec/2_related_work}

\input{sec/3_method}
\input{sec/4_benchmarks}
\input{figure_latex/aiced_summary_300}

\input{sec/5_experiments}
\input{sec/6_disscussion_limitations}

\input{sec/7_conclusion}
\clearpage


\appendix


\input{sec/X_suppl}

\clearpage
%
%
\bibliographystyle{splncs04}
\bibliography{main}
\end{document}

%% file: sec/0_abstract.tex
\begin{abstract}
Text-to-image diffusion models suffer from the risk of generating outdated, copyrighted, incorrect, and biased content. While previous methods have mitigated the issues on a small scale, it is essential to handle them simultaneously in larger-scale real-world scenarios. We propose a two-stage method, Editing Massive Concepts In Diffusion Models~(EMCID). The first stage performs memory optimization for each individual concept with dual self-distillation from text alignment loss and diffusion noise prediction loss. The second stage conducts massive concept editing with multi-layer, closed form model editing.
We further propose a comprehensive benchmark, named ImageNet Concept Editing Benchmark~(ICEB), for evaluating massive concept editing for T2I models with two subtasks, free-form prompts, massive concept categories, and extensive evaluation metrics.
Extensive experiments conducted on our proposed benchmark and previous benchmarks demonstrate the superior scalability of EMCID for editing up to 1,000 concepts, providing a practical approach for fast adjustment and re-deployment of T2I diffusion models in real-world applications.
\keywords{T2I Generation, Diffusion Model, Concept Editing}

\end{abstract}

%% file: sec/1_intro.tex
\section{Introduction}
\label{sec:intro}

Text-to-image diffusion models~\cite{dhariwal2021diffusion,rombach2022high,sdv1.4, ho2022cascaded, ramesh2022hierarchical, Dall-e-3, Midjourny, saharia2022photorealistic} have advanced remarkably in recent years. However, various societal concerns have also been raised~\cite{luccioni2023stable, struppek2022biased, somepalli2023diffusion, shan2023glaze}. These models may produce inaccurate content due to outdated or flawed internal knowledge. They may also pose risks related to copyright infringement and societal biases inherited from training data. 
While the inappropriate generation largely stems from related data in the unfiltered web-scale training set, it is prohibitively expensive to resolve the issues by reprocessing the training set and retraining the models. To provide more practical solutions, we would edit models' knowledge of concepts related to the issues, by modifying a portion of the model weights.

Previous methods either fine-tune the T2I model~\cite{kumari2023ablating, gandikota2023erasing, kim2023towards, heng2023selective} or adopt existing approaches from editing large language models~\cite{orgad2023editing, arad2023refact, gandikota2023unified}.
However, most of them modify model weights sequentially when editing multiple concepts, leading to the \textit{catastrophic forgetting}~\cite{mccloskey1989catastrophic} phenomenon where the model degenerates with the increasing number of edits. 
Recent work UCE~\cite{gandikota2023unified} unifies multiple concept editing scenarios and edits multiple concepts in parallel.
However, it still faces a huge generation quality drop when editing more than 100 concepts.
So a challenge arises: \textit{How to preserve the generation ability of a T2I model when massive concept editing such as editing over 1,000 concepts?}

To tackle this challenge, we propose a two-stage framework, \textit{Editing Massive Concepts In Diffusion Models}~(EMCID).
The first stage performs decentralized memory optimization of each individual concept with dual self-distillation loss. The dual self-distillation loss aligns both the text features of the text encoder and the noise predictions of the U-Net, encouraging semantics-aware and visual-detail-aware accurate memory optimization of individual concepts.
In the second stage, the optimization for individual concepts are aggregated for editing in parallel. We derive multi-layer closed-form model editing to enable massive concept editing. 
Our designs enable \method{} to excel in large-scale concept editing in T2I models, allowing successful editing of up to 1,000 concepts.  

To conduct comprehensive evaluations and analysis of concept editing methods, we curate a new benchmark, named ImageNet Concept Editing Benchmark~(ICEB).
In addition to the sub-task of large-scale arbitrary concept editing (including concept updating and concept erasing), we further propose a novel and applicable sub-task, Concept Rectification, which is to rectify the incorrect generation results of the less popular aliases of concepts.
In contrast to earlier benchmarks, which may suffer from small-scale evaluations, imprecise metrics, or limited evaluation prompts, our benchmark is equipped with free-form prompts, up to 300 concept edits, and extensive metrics.

In summary, our contributions are three-fold. (1) We propose a two-stage pipeline, EMCID, to edit T2I diffusion models with high generalization ability, specificity, and capacity. The dual self-distillation supervision in stage I forces the model to be aware of both semantics and visual details of the editing concept. The multi-layer editing and closed-form solution in stage II enable massive concept editing. 
(2) We create a comprehensive benchmark to evaluate concept editing methods for T2I diffusion models, spanning a magnitude of up to 300 edits with two sub-tasks, free-form prompts, and extensive evaluation metrics. 
(3) Extensive experiments demonstrate the scalability of EMCID in editing massive concepts (up to 1,000 concepts), surpassing previous approaches that can only edit at most 100 concepts.

%% file: sec/2_related_work.tex
\section{Related Work}\label{sec:related work}
\textbf{Text-to-image diffusion models.}
Diffusion models have been successfully applied for text-to-image generation~\cite{sdv1.4,rombach2022high,saharia2022photorealistic,mou2023t2iadapter,Dall-e-3,Midjourny}.
With the advanced T2I diffusion models gaining more popularity, they give rise to risks caused by various issues: generation of images reflecting outdated or incorrect knowledge, copyright infringement, and reinforcement of societal biases~\cite{cho2023dalleval,luccioni2023stable}. These problems can be mitigated through extensive preparation and modification~\cite{9093454,sdv2.0} of training data. However, this approach demands a significant investment of time and computational resources. Therefore, the importance of a method that can handle diverse concept editing tasks and enable extensive-scale editing cannot be overstated. Our approach meets this need by scaling up the capacity for edited concepts to a remarkable 1,000.

\noindent \textbf{Fine-tuning T2I models for concept editing.}
A line of previous methods~\cite{kumari2023ablating,gandikota2023erasing,kim2023towards,heng2023selective,zhang2023forgetmenot} fine-tune T2I diffusion models, particularly the cross-attention layers, to selectively edit source concepts. A part of them focuses on erasing concepts~\cite{zhang2023forgetmenot,gandikota2023erasing, kim2023towards} while the others~\cite{kumari2023ablating, heng2023selective} generally edits source concepts as destination concepts.    
However, during the continuous fine-tuning to edit multiple concepts, these methods often encounter issues including catastrophic forgetting and significant time costs. Our method does not fine-tune model weights directly, instead, we edit the weights with closed-form solutions. 

\noindent \textbf{Concept editing with closed-form solutions.} To edit concepts in T2I diffusion models, model-editing-based methods~\cite{arad2023refact,gandikota2023unified, orgad2023editing} are another line of work, modifying a model's weights with closed-form solutions. These methods take inspiration from the success of knowledge-editing in NLP, where~\cite{meng2022locating,Meng2022memit} have introduced the perspective of viewing MLPs as linear associative memories~\cite{anderson1972simple, kohonen1972correlation} and successfully edited knowledge within LLMs from this perspective. Among editing based methods for T2I diffusion models, ReFACT~\cite{arad2023refact} takes inspiration from the method of ROME~\cite{meng2022locating} and edits the text encoder of Stable Diffusion~\cite{rombach2022high}, while UCE~\cite{gandikota2023unified} and TIME~\cite{orgad2023editing} edit the cross-attention layers. Our method is also a model-editing-based method, while being able to edit a much larger number of concepts, compared to previous methods. Different from previous methods, our method gives attention to the diffusion process itself and meticulously designs the approaches to edit the text encoder of T2I diffusion models.

%% file: sec/3_method.tex
\section{Method}
\label{sec:method}

\vspace{-0.10in}
\subsection{Overview}
\label{sec:preliminary}

\noindent
\textbf{Task formulation.}
We integrate various types of concept editing tasks for text-to-image generation, including updating concepts, erasing art styles, rectifying imprecise generation, and gender debias, into a unified formulation.
We define concept editing in text-to-image generation as modifying the generated images conditioned on the source concept to match the destination concept.
This problem formulation unifies various types of concept editing tasks, as shown in Tab.~\ref{tab:problem setup}.
\input{tables/problem_setup}

\noindent\textbf{Where to edit?}
We consider the text-to-image diffusion models composed of a transformer-based text encoder $E(p)$ that encodes the input text prompt $p$ into feature embeddings and the U-Net image generator that predicts the noise maps $\boldsymbol{\epsilon}$ conditioned on the text embeddings.
It is intuitive to assume that the most textual and semantic knowledge are stored in the text encoder $E$ and the image prior are stored in the U-Net generator. In addition, prior works~\cite{meng2022locating,Meng2022memit} revealed that the feed-forward multi-layer perceptions (MLPs) store the factual knowledge in large language models.
\textit{Therefore, we focus on the MLP layers in the text encoder in order to edit concepts in the text-to-image diffusion models.}
Compared with previous approaches that edit cross-attention layers~\cite{orgad2023editing, gandikota2023unified} or finetune U-Net parameters~\cite{kumari2023ablating, gandikota2023erasing}, our approach ensures large concept editing capacity without affecting the image generation quality.

\noindent\textbf{How to edit?}
Each MLP in transformer consists of two weight matrices with a non-linear activation in between, formulated as $W_{proj}\cdot \sigma(W_{fc})$. We view $W_{proj}$ as a linear associative memory~\cite{anderson1972simple,kohonen1972correlation}, following previous work~\cite{meng2022locating,Meng2022memit}. From this perspective, a linear projection is a key-value store $WK\approx V$, which associates a set of input keys $K_0=[k_1\mid k_2 \mid \cdots k_n]$ to a set of corresponding memory values $V_0=[v_1 \mid v_2 \mid \cdots v_n]$.
Therefore, the goal of our model editing is to add new key-value pairs, $K_1=[ k_{n+1} \mid k_{n+2} \mid \cdots \mid k_{n+e}]$ and $V^*_1=[ v^*_{n+1} \mid v^*_{n+2} \mid \cdots \mid v^*_{n+e}]$, into the associative memories while preserving existing key-value associations.
Mathematically, we formulate the objective as:
\begin{equation}
\centering
\label{eq:model editing objective}
    W^*=\underset{W}{\text{argmin}}
    ((1 - \alpha)\sum\limits_{i=1}^{n}||Wk_{i} - v_{i}||^{2} + \alpha\sum\limits_{i=n+1}^{n+e}||Wk_{i} - v^*_{i}||^{2}
)
\end{equation}
where $\alpha$ is a hyperparameter to control the trade-off between preserving existing memories and editing new concepts.
The existing key-value pairs $K_0$ and $V_0$ are estimated on the large-scale image-caption pairs (In practice, we use CCS (filtered) dataset from BLIP~\cite{li2022blip}).
The key vectors in $K_1$ representing the source concepts are derived from the features of the last subject token in the source prompt, following previous work~\cite{meng2022locating, Meng2022memit, arad2023refact}.
The remaining problems are how to derive $V^*_1=[ v^*_{n+1} \mid v^*_{n+2} \mid \cdots \mid v^*_{n+e}]$ and how to solve the overall optimization objective.

\noindent\textbf{Overview of EMCID.}
Our \method{} is a two-stage method that edits multiple layers of the text encoder of T2I diffusion models with closed-form solutions, as illustrated in Fig.~\ref{fig:ecid framework}.
The first stage (Sec.~\ref{sec:ecid stage1}) performs decentralized memory optimization for each individual concept, aiming to optimize $v^*_i$ for each $k_i$ in Eq.~\ref{eq:model editing objective}. With the proposed dual self-distillation optimization, the differences between the source concept and the destination concept are distilled into an optimized feature offset vector, 
from both semantic concepts of the text encoder and visual concepts of the diffusion model.
The second stage (Sec.~\ref{sec:ecid stage2}) aggregates the optimized $v^*_i$ from individual concept optimization in the first stage to optimize the objective in Eq.~\ref{eq:model editing objective}. We perform multi-layer model editing with closed-form solutions to enable massive concept editing.

\input{figure_latex/ecid_framework}

\subsection{Stage I: Memory Optimization with Dual Self-Distillation}
\label{sec:ecid stage1}

The goal of this stage is to obtain the value vectors $v^*$ for each key of the concept to edit, which will be further used for the second term of the objective in Eq.~\ref{eq:model editing objective}.

Specifically, given a group of source prompts $p$ and destination prompts $\hat{p}$, we encode the source text prompts $p$ with the transformer-based text encoder, and compute the average value of the last subject token after the non-linear activation in the $l$-th MLP as the key $k$. The original value associated with the source concept is $v=Wk$, and we aim to optimize an offset vector $\delta^*$ so that the new value $v^*=v+\delta^*$ can associate the source concept to the destination concept.

In order to capture both the semantic-level concept and the visual details of the destination concept, we design a novel dual self-distillation method to optimize $\delta^*$ with text alignment loss from the text encoder and noise prediction loss from the diffusion model.

\noindent\textbf{Self-distillation of semantic concepts from text encoder.}
In order to associate the source concept with the destination concept, we optimize $\delta$ with the text alignment loss $\mathcal{L}_\text{txt}$. It aligns between the source prompt embedding with updated values and the destination prompt embedding, as the following optimization objective:
\begin{equation}
\centering
\label{eq:txt_loss}
\delta^*=\underset{\delta}{\text{argmin}}||E_{v=+\delta}(p)-E(\hat{p})||^2,
\end{equation}
where $E(\cdot)$ denotes the feature vector of the [EOS] token encoded by the text encoder, representing the embedding of the whole text prompt. The notation $E_{v=+\delta}$ represents that we modify the computation of the text encoder by substituting the value vector $v$ of the last subject token with $v+\delta$. 

\noindent\textbf{Self-distillation of visual concepts from diffusion model.}
The self-distillation from text encoder only considers the text feature alignment between the source prompt and the target prompt, but ignores the information from the diffusion-based image generation model.
Only aligning the text embeddings does not guarantee our final goal of generating images with destination concepts.
Therefore, we propose the noise prediction loss $\mathcal{L}_\text{noise}$ to distill visual knowledge of the destination concepts from diffusion models.
The optimization objective is:
\begin{equation}
\centering
\label{eq:noise_loss}
\delta^*=\underset{\delta}{\text{argmin}}\mathbb{E}_{\mathbf{x}_{t},t}||\boldsymbol{\epsilon}(\mathbf{x}_{t}, E_{v=+\delta}(p), t) - \boldsymbol{\epsilon} (\mathbf{x}_{t},E(\hat{p}),t)||^2,
\end{equation}
where $\mathbf{x}_{t}$ denotes the images generated from the destination prompts added by noise with timestep $t$, $E(\cdot)$ denotes the text embeddings which are injected to the diffusion U-Net with cross-attention layers, $\boldsymbol{\epsilon} (\mathbf{x}_{t},E(\hat{p}),t)$ represents the noise prediction with the destination prompt, and $\boldsymbol{\epsilon}(\mathbf{x}_{t}, E_{v=+\delta}(p), t)$ represents the noise prediction from the source prompt with the optimized offset $\delta$. The noise prediction loss enables end-to-end optimization of $\delta$ with a direct constraint, that the images generated from the source prompts should match the images generated from the destination prompts. In addition, the self-distillation from diffusion models allows the users to provide destination images instead of destination text prompts, in scenarios where the destination concepts are difficult to describe with text prompts. In this case, the optimization objective is:
\begin{equation}
\centering
\label{eq:noise_loss_2}
\delta^*=\underset{\delta}{\text{argmin}}\mathbb{E}_{\mathbf{x}_{t},t}||\boldsymbol{\epsilon}(\mathbf{x}_{t}, E_{v=+\delta}(p), t) - \boldsymbol{\epsilon}_t||^2,
\end{equation}
where $\mathbf{x}_{t}$ denotes the user-provided images added by noise with timestep $t$, $\boldsymbol{\epsilon}_t$ denotes the ground-truth noise added to the user-provided image, and other notations remain the same with Eq.~\ref{eq:noise_loss}.

\noindent\textbf{Dual self-distillation.}
We derive the overall optimizing objective by integrating the self-distillation objective $\mathcal{L}_{\text{txt}}$ from the text encoder (Eq.~\ref{eq:txt_loss}) and the self-distillation objective $\mathcal{L}_{\text{noise}}$ from the diffusion models (Eq.~\ref{eq:noise_loss} or Eq.~\ref{eq:noise_loss_2}). The dual self-distillation enables us to find the optimal $\delta^*$ and derive the updated value $v^*=v+\delta^*$ to associate the source concept with the destination concept.

\subsection{Stage II: Model Editing for Massive Concepts}
\label{sec:ecid stage2}

\noindent\textbf{Closed-form model editing for massive concepts.}
The previous value optimization stage finds the optimal values $V^*_1=[ v^*_{n+1} \mid v^*_{n+2} \mid \cdots \mid v^*_{n+e}]$ for each concept to edit. Back to our final goal of editing massive concepts and the overall objective of modifying the weight matrix $W$ to minimize the objective function in Eq.~\ref{eq:model editing objective}. Following previous work~\cite{Meng2022memit}, we derive the closed-form solution for the editing objective as, 

\begin{equation}
\centering
\label{eq:closed-form solution}
     W^* = W_0 + 
\alpha (V^*_1 - W_0 K_1)K_{1}^{T}
\left[(1 - \alpha)K_0 K_{0}^T + \alpha K_{1}K_{1}^{T} \right]^{-1},
\end{equation}
where $W_0$ is the original weight matrix. The editing intensity hyperparameter $\alpha$ controls the trade-off between editing concepts and preserving existing knowledge, and is set to 0.5 by default.
Detailed math derivation and experiments exploring the trade-off effect concerning $\alpha$ are included in the Appendix Sec.~\ref{sec:ablation}.

\noindent\textbf{Multi-layer model editing for massive concepts.}
Editing massive concepts requires a large capacity of the model parameters that can be updated. Therefore, instead of editing a single MLP layer, we propose to update multiple MLP layers in the text encoder.
Specifically, we sequentially edit the weight matrices of the MLP layers from the shallow layers to the deep layers. 
It is worth mentioning that previous work MEMIT~\cite{Meng2022memit} observes improved robustness when spreading the weight updates to multiple layers in large language models, which evidences our design from another perspective.
Different from MEMIT which spreads the weight updates over the critical path, we conduct ablation studies and analysis on which layers to spread the weight updates over and how the selection of layers will affect the concept editing performance in Sec.~\ref{sec:exp ablation}.

%% file: tables/problem_setup.tex
\newcolumntype{C}[1]{>{\centering\arraybackslash}p{#1}}

\begin{table}
\vspace{-0.25in}
\caption{Our problem setup for various tasks. We give an example for each task.}
\vspace{-0.1in}
\centering
\fontsize{6pt}{8pt}\selectfont
\begin{tabular}{C{2.5cm}C{3cm}C{3cm}C{3cm}}
\toprule
\textbf{Tasks} & \textbf{Examples} & \textbf{Source Prompts} & \textbf{Destination Prompts}  \\
\midrule
Updating Concepts & Update US president as Joe Biden & ``US president'' & ``Joe Biden'' \\
\midrule
Erasing Art Styles & Van Gogh to normal style & ``Image in Van Gogh style'' & ``Image in normal art style'' \\
\midrule
Rectify Imprecise Generation & Rectify snowbird generation & ``Snowbird'' & ``Junco''~(a more popular name)\\
\midrule
Gender Debias & Balance "doctor" gender ratio & ``Doctor''& ``Female doctor''/``Male doctor''~(1:1) \\
\bottomrule
\end{tabular}

\label{tab:problem setup}
\vspace{-0.2in}
\end{table}


%% file: figure_latex/ecid_framework.tex
\begin{figure*}
\vspace{-1em}
    \centering
\includegraphics[width=\linewidth]{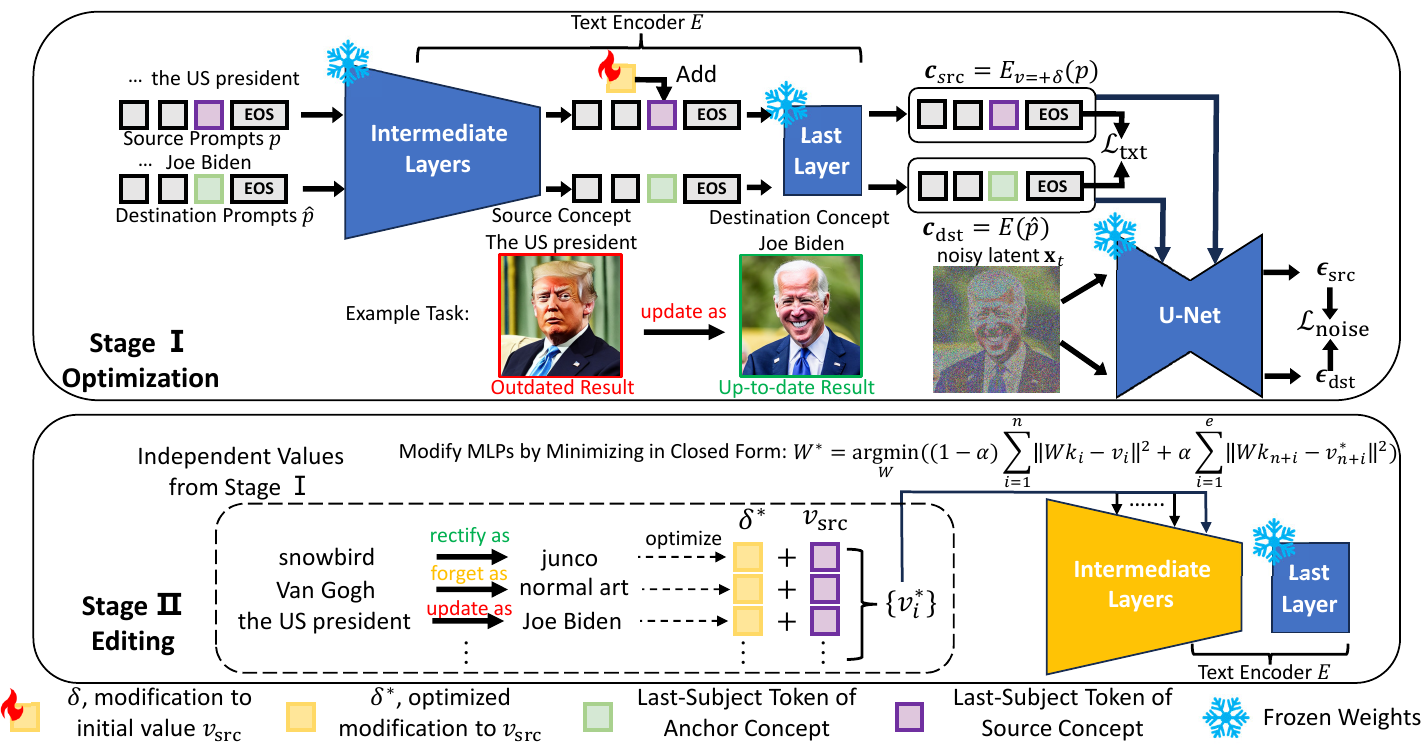}
    \caption{
    The two-stage pipeline of EMCID. We demonstrate stage I with the example of updating the source concept, ``the US president'', as the destination concept ``Joe Biden''. In the first stage, we align both the embeddings of the text prompts and the noise predictions $\boldsymbol{\epsilon}_{\text{dst}}\triangleq \boldsymbol{\epsilon}(\mathbf{x}_{t},
      \mathbf{c}_{\text{dst}},t)$ and $\boldsymbol{\epsilon}_{\text{src}}\triangleq \boldsymbol{\epsilon}(\mathbf{x}_{t}, \mathbf{c}_{\text{src}}, t)$. Multiple source concepts can be independently updated. 
      In stage II, we edit the MLPs of the intermediate layers of the text encoder using a closed-form solution based on the independent values obtained from stage I. 
      } 
      \vspace{-0.25in}
    \label{fig:ecid framework}
    
\end{figure*}


%% file: sec/4_benchmarks.tex
\vspace{-0.10in}
\section{Benchmark}
\label{sec:benchmark}
\input{tables/benchmark_cmp}
In previous literature, there has been a notable lack of a standard benchmark capable of evaluating concept editing methods for T2I diffusion models at a large scale. To address this problem, we have curated a comprehensive benchmark, ImageNet Concept Editing Benchmark~(ICEB). Our benchmark consists of two sub-tasks. The first sub-task is designed for general evaluations in terms of the ability to update or erase arbitrary concepts, allowing for updating or erasing up to 300 concepts. The second sub-task aims at a novel and practical application, specifically rectifying the incorrect generation results conditioned on less popular aliases of concepts.
In contrast to indirect metrics built on CLIP accuracy in previous work~\cite{kumari2023ablating, orgad2023editing, arad2023refact}, we propose comprehensive metrics to evaluate a concept editing method from diverse aspects, as shown in Fig.~\ref{fig:aiced_300}.

\vspace{-0.10in}
\subsection{Data Collection}
Utilizing ChatGPT~\cite{ChatGPT}, we obtained 3,300 diverse and effective prompts for 666 classes from ImageNet~\cite{deng2009imagenet} after filtering prompts and classes that struggle to guide the generation of correct images. 
For filtering, we utilized ViT-B~\cite{dosovitskiy2020image} as the evaluator.  Only those prompts were retained whose generated images attained a classification probability exceeding 0.5, as determined by ViT-B, accurately reflecting the classes they describe. To present the potential editing effect gap between simple template prompts and more diverse prompts, We also collected 900 template-based prompts for evaluation. More details are in Appendix Sec.~\ref{sec:data collection}.

\vspace{-0.10in}
\subsection{Task definition}
\label{sec:task definition}
\noindent\textbf{Arbitrary Concept Editing.}
We define the first sub-task as a general task, Arbitrary Concept Editing, where a set of source concepts from ImageNet classes are updated as another set of destination concepts from ImageNet. This task supports updating up to 300 concepts. Concretely, we randomly sample 300 classes from the 666 collected classes of ICEB as source concepts. For each source concept, a destination concept is sampled from the 5-nearest concepts measured by CLIP text distance among the 366 classes left. 
We measure the performance of concept editing methods on this task by the success of editing, the generalization to various aliases and prompts, and the preservation of non-source concepts. Detailed definitions of these metrics are demonstrated in Sec.~\ref{sec:metric}.
Moreover, the ability of erasing concepts can also be evaluated on this task by setting the destination concepts as null.

\noindent\textbf{Concept Rectification.}
We test the performance of Stable Diffusion v1.4 on the less popular aliases on ImageNet, and we observe that many aliases cannot guide the model to generate correct images, as shown in Fig.~\ref{fig:imgnet_mend_cmp}. This phenomenon gives rise to the demand for rectifying the generation results of T2I models. 
Based on the observation, we propose a novel sub-task for ICEB, named Concept Rectification. The task is to rectify the incorrect generation of ``misunderstood aliases'' of ImageNet classes. Concretely, we collected 140 misunderstood aliases and 700 evaluation prompts based on ICEB. This task is evaluated by the extent of rectifying these concepts and preserving existing knowledge.

\vspace{-0.15in}
\subsection{Evaluation Metrics}
\label{sec:metric}
\input{tables/evaluation_metrics}

In previous benchmarks~\cite{orgad2023editing, arad2023refact}, the success of an edit was determined by assessing if the generated images conditioned on the source concept resemble the destination concept more than the source concept itself, as evaluated by CLIP~\cite{radford2021clip}. However, this approach overlooks the degree to which the source concepts have been altered to resemble the destination concepts.
Based on this observation, we propose four novel metrics, as listed in Tab.~\ref{table:evaluation_metrics}. 

For Source Forget and Source2Dest, template prompts (\eg ``an image of \{\}'') and diverse prompts~(generated by ChatGPT) are used separately to evaluate the editing \textit{efficacy} and \textit{generalization} ability towards more complex prompts. For other metrics we use ChatGPT generated diverse prompts for calculation.

Specifically, we define $p_{M}(a, b)$ as the average confidence that the images generated by the T2I model $M$ conditioned on class $b$ are classified as class $a$, by a ViT-B image classification model pretrained on ImageNet.
We denote the original T2I model by $M$ and the edited model by $\hat{M}$. 
Thus the metrics are defined as 
in the col.~3, Tab.~\ref{table:evaluation_metrics}.
Here $S$, $AL$, and $H$ represent the numbers of edited source concepts, aliases, and non-edited holdout concepts, respectively. $s_i$, $al_i$, $d_i$, and $h_i$ denote the $i$-th edited source concept, alias, destination concept, and non-edited concept, respectively.

For Concept Rectification, the source concepts are the misunderstood aliases, and the destination concepts are defined as their classes. So, Source2Dest is used to measure the success of rectifying the generation results of the aliases. We don't calculate Source Forget or Alias2Dest because they are ill-defined for this task.
We further use CLIP score and FID~\cite{fid} on COCO-30k prompts~\cite{lin2014microsoft} to evaluate the preservation of the T2I model's generation capabilities.

%% file: tables/benchmark_cmp.tex
\newcolumntype{C}[1]{>{\centering\arraybackslash}p{#1}}

\begin{table}
\vspace{-0.3in}
\caption{
Comparison of concept editing benchmarks.
 }
 \vspace{-0.1in}
\centering
\fontsize{7pt}{9pt}\selectfont
\begin{tabular}{C{2.5cm}C{3cm}C{1.5cm}C{1.5cm}C{1.5cm}C{2cm}}
\toprule
\textbf{Benchmark}  &
\textbf{Prompt Diversity} & \textbf{Prompts} &\textbf{Concepts} & \textbf{Metrics} & \textbf{Tasks}\\
\midrule
TIMED\cite{orgad2023editing} &Template & 410 & 82 & 4 &Update \\
RoAD\cite{arad2023refact} &Template & 450 & 90 & 4 &Update \\
Artists-Forget\cite{gandikota2023unified}& Template & 7500 & 1000 & 4 & Forget\\
Gender-Debias\cite{gandikota2023unified} &Template & 175 & 35 & 3 & Debias \\
\midrule
ICEB &ChatGPT+Template & 3330+900 & 300& 6 & {Update,Rectify}\\
\bottomrule
\end{tabular}
 \vspace{-0.3in}
\end{table}


%% file: tables/evaluation_metrics.tex
\newcolumntype{C}[1]{>{\centering\arraybackslash}m{#1}}

\begin{table}
\vspace{-0.1in}
\centering
\caption{Summary of evaluation metrics}
\vspace{-0.1in}
\label{table:evaluation_metrics}
\fontsize{8pt}{10pt}\selectfont
\begin{tabular}{C{2.2cm}C{4.5cm}C{5.8cm}}
\toprule
\textbf{Metric Names} & \textbf{Meanings of the Metrics} & \textbf{Definition Equations} \\
\midrule
\makecell{Source Forget \\ (\textbf{SF})} 
& The effectiveness of forgetting original source concepts. 
& \( \text{SF} = 
    \frac{1}{S} 
    \sum\limits_{i=1}^{S}
        \left[p_{M}(s_i, s_i) - p_{\hat{M}}(s_i, s_i)\right]\) \\

\midrule
\makecell{Source2Dest \\ (\textbf{S2D})}
& The effectiveness of transforming source concepts into destination concepts. 
& \( \text{S2D} = 
    \frac{1}{S} 
    \sum\limits_{i=1}^{S}
        \left[p_{\hat{M}}(s_i, d_i) - p_{M}(s_i, d_i)\right] \) \\
\midrule
\makecell{Alias2Dest \\ (\textbf{AL2D})} 
& The effectiveness of transforming the aliases of the source concepts into destination concepts. 
& \( \text{AL2D} = 
    \frac{1}{AL} 
    \sum\limits_{i=1}^{AL}
        \left[p_{\hat{M}}(al_i, d_i) - p_{M}(al_i, d_i)\right] \) \\
\midrule
\makecell{Holdout Delta \\ (\textbf{HD})} 
& The drop in generation capabilities for non-edited holdout concepts caused by the edits. 
& \( \text{HD} = 
    \frac{1}{H} 
    \sum\limits_{i=1}^{H}
        \left[p_{\hat{M}}(h_i, h_i) - p_{M}(h_i, h_i)\right] \) \\
\bottomrule
\end{tabular}
\vspace{-0.2in}
\end{table}


%% file: figure_latex/aiced_summary_300.tex
\begin{figure*}[t]
\vspace{-0.1in}
    \centering
    \includegraphics[width=\linewidth]{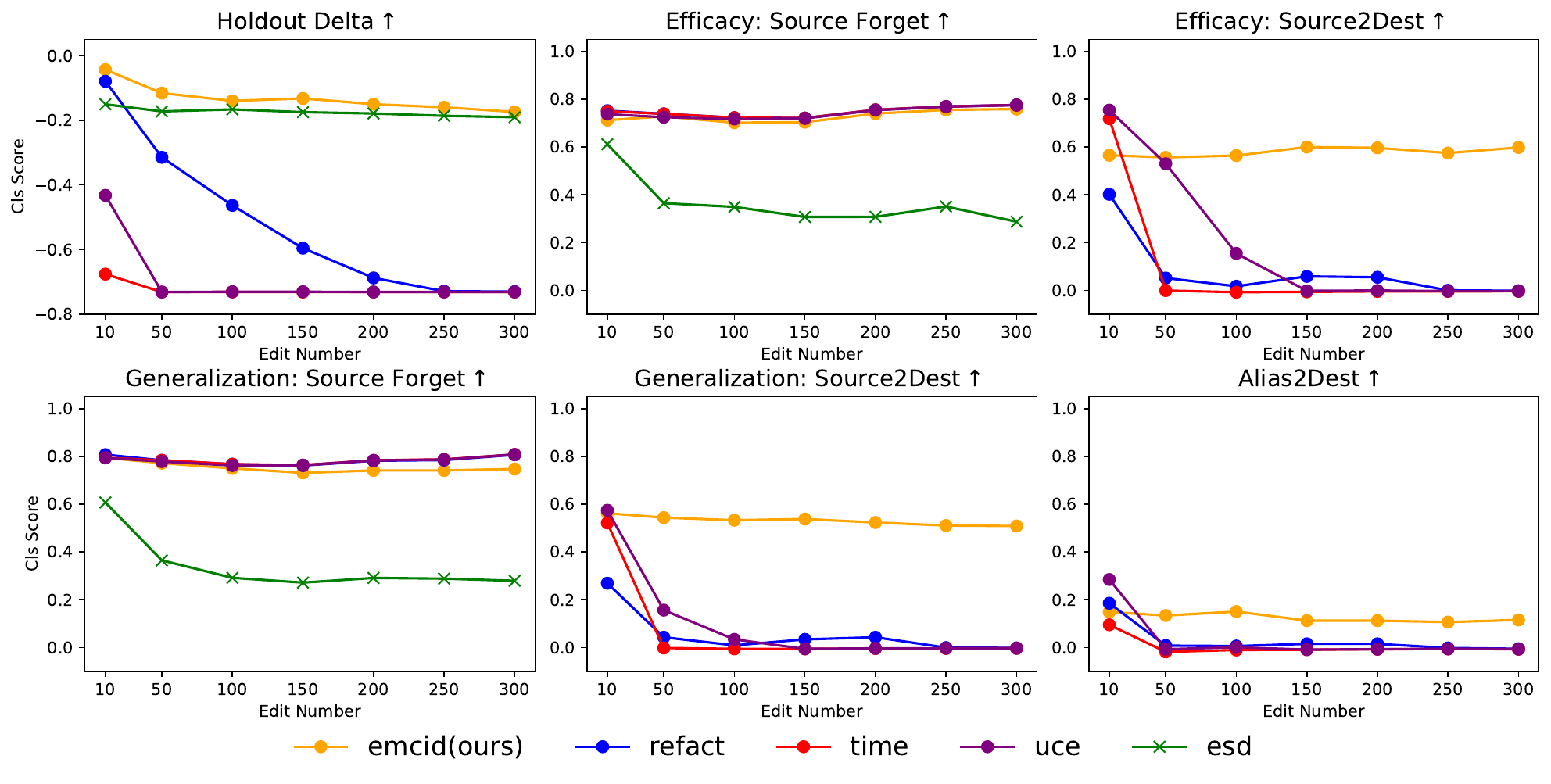}
    \vspace{-0.30in}
    \caption{
    We present comparisons on the task of Arbitrary Concept Editing. We use the dot marker for methods editing source concepts as designated concepts, and the cross marker for concept erasing methods. Source2Dest and Alias2Dest are not suitable for comparisons for concept erasing methods, thus not presented for them. Our \method{} can successfully edit up to 300 concepts with minor influence to holdout concepts. In comparison, the success of our baselines in editing source concepts and preserving holdout concepts exhibits a rapid decline as the number of edits increases.
    }
    \vspace{-0.3in}
    \label{fig:aiced_300}
\end{figure*}

%% file: sec/5_experiments.tex
\vspace{-0.15in}
\section{Experiments}
\label{sec:experiments}
In this section, we first demonstrate experiments setup in Sec.~\ref{sec:exp setup}, then we compare our method with both fine-tuning-based and model-
editing-based baselines on the two sub-tasks of ICEB, in Sec.~\ref{sec:exp ACE} and Sec.~\ref{sec:exp CR}. We also compare our method on the task of erasing 1,000 artist styles with the recent SOTA method UCE~\cite{gandikota2023unified} in Sec.~\ref{sec:exp art}. We further present ablation studies about our method in Sec.~\ref{sec:exp ablation}. Our method achieves comparable results to UCE in gender-debiasing multiple professions, as explained in the Appendix Sec.\ref{sec:debias}. 
\vspace{-0.10in}
\subsection{Experiments Setup}
\label{sec:exp setup}
We test our method and baselines for concept editing tasks on the SD v1.4~\cite{sdv1.4} model. By default, we utilize dual self-distillation in stage I, integrating objectives from Eq.~\ref{eq:txt_loss} and Eq.~\ref{eq:noise_loss}, as a hybrid loss $\mathcal{L}_\text{hybrid}$, with a balance factor $\lambda_s$ set empirically as 0.01:
\begin{equation}
\centering
\label{eq:hybrid_loss}
\delta^*=\underset{\delta}{\text{argmin}}(\mathcal{L}_\text{noise} + \lambda_s \mathcal{L}_\text{txt})
\end{equation}
This optimization stage for each concept takes only 200 gradient update steps with a constant $0.2$ learning rate. For stage II, by default we spread the weight updates over all the layers of the text encoder except the last one, which can largely enhance the editing effect, according to our ablation studies in Sec.~\ref{sec:exp ablation}. On an NVIDIA RTX 4090 GPU, stage I for each concept can be accomplished within 92 seconds, and stage II for aggregating eidts of 300 concepts can be completed within 10 seconds. The parallelization of the first stage and the fast speed of the second stage allow our method to edit massive concepts in a significantly shorter period, in contrast to methods sequentially editing the concepts. For prompts $p$ and $\hat{p}$ used for training, we chose simple template-based prompts, as a typical approach of previous methods~\cite{gandikota2023unified, orgad2023editing}, for fair comparisons.

\vspace{-0.10in}
\subsection{Large-Scale Arbitrary Concept Editing}
\label{sec:exp ACE}
\noindent
\textbf{Setup.}
In this experiment, we employ concept editing methods for T2I diffusion models on the Arbitrary Concepts Editing task, as defined in Sec.~\ref{sec:task definition}. We compare with methods capable of editing more than 10 concepts on ICEB. Among them, ESD-x~\cite{gandikota2023erasing} fine-tunes the cross-attention layers of Stable Diffusion to erase concepts. Besides, UCE~\cite{gandikota2023unified}, TIME~\cite{orgad2023editing} and ReFACT~\cite{arad2023refact} modify either cross-attention layers or the text encoder with closed-form solutions, to alter source concepts towards destination concepts. We set $\alpha=0.6$ for \method{} on this task.

\noindent
\textbf{Metrics.}
We use metrics defined in Sec.~\ref{sec:metric} for evaluation. For concept erasing methods, we only present 2 metrics, Source Forget and Holdout Delta, as the remaining metrics are unsuitable for comparisons in their case.

\noindent
\textbf{Analysis.}
As shown in Fig.~\ref{fig:aiced_300}, \method{} can edit the source concepts as destination concepts with high Source2Dest across different edit scales, while for all baselines the success of edits drops quickly with the increase of edit number. Moreover, \method{} presents superior specificity for preserving non-source concepts, proved by the relatively minor Holdout Delta. The edit effect of \method{} can also generalize to diverse prompts and aliases of the source concepts.

\input{figure_latex/imgnet_mend_cmp}
\input{tables/ICMD_results}

\input{figure_latex/artists_holdout_cmp}
\vspace{-0.25in}
\subsection{Concept Rectification}
\vspace{-0.02in}
\label{sec:exp CR}
\input{figure_latex/rectify_class}
\noindent
\textbf{Setup.}
We compare \method{} with the SOTA method UCE~\cite{gandikota2023unified} on the task of rectifying 140 misunderstood aliases of concepts, defined as Concept Rectification in Sec.~\ref{sec:task definition}. We also conduct qualitative experiments to rectify 8 misunderstood aliases, as shown in Fig.~\ref{fig:imgnet_mend_cmp}. The rectification is accomplished by editing the misunderstood aliases~(\eg, ``Snowbird'') as its popular class name~(\eg, ``Junco'' ) which can correctly guide the generation of the T2I diffusion model. Because UCE needs to specify the concepts to preserve, we choose 200 ImageNet classes as concepts for it to preserve. We further apply \method{} to rectify 6 classes that cannot be generated properly by Stable Diffusion v1.4, using only reference images from ImageNet validation set. Note that for this scenario, previous model editing methods~\cite{arad2023refact, orgad2023editing, gandikota2023unified}, including UCE, are not applicable.

\noindent
\textbf{Metrics.}
As explained in Sec.~\ref{sec:metric}, for large-scale quantitative evaluation, we use Source2Dest to measure the success of the edits. And for evaluating the preservation of generation capabilities, besides Holdout Delta we further measure CLIP score and FID on COCO-30k prompts.

\noindent
\textbf{Analysis.}
As shown in Fig.~\ref{fig:imgnet_mend_cmp}, UCE struggles to rectify aliases even at a small scale, while our method can effectively correct the results. We demonstrate that this is because UCE's objective neglects the diffusion process and only focuses on the encoding of the texts, while our method maintains high efficacy through dual self-distillation. What's more, when only reference images are available, our method can still successfully rectify the source concepts, as shown in Fig.~\ref{fig:rectify_class}.

\input{figure_latex/artists_edit_cmp}

\subsection{Erasing Artist Styles}
\label{sec:exp art}
\noindent
\textbf{Setup.}
For a fair comparison, we follow the experiment setting of UCE. We conduct experiments for erasing from 1 to 1,000 artist styles, and erase an artist's style~(\eg ``Van Gogh'') by editing it as ``art''. Note that it is also feasible to edit the styles as any designated concepts with our method. We present qualitative results of our \method{} in Fig.~\ref{fig:artists_edit_cmp}, showcasing the effect of erasing up to 1,000 artist styles. We further compare with UCE, for the preservation of both 500 non-edited artist styles and the model's overall generation capabilities, as shown in Fig.~\ref{fig:artists_holdout}. We observe that erasing artist styles is a less challenging task and requires less parameter modifications, so we edit the 7 to 10 instead of 0 to 10~(our default setting) layers of the text encoder, based on observations in Sec.~\ref{sec:exp ablation}.

\noindent
\textbf{Metrics.}
To evaluate the preservation of non-source artist styles, we calculate the CLIP score of the modified T2I model on 2,500 prompts trying to mimic the 500 holdout artist styles. LPIPS~\cite{zhang2018unreasonable} between images generated by pre-edit and post-edit T2I models is also used to measure the influence on holdout artist styles. A higher CLIP score or lower LPIPS means better preservation of holdout styles. For the preservation of overall generation capabilities, we adopt the typical approach of measuring CLIP score and FID on COCO-30k prompts.

\noindent
\textbf{Analysis.}
As shown in Fig.~\ref{fig:artists_edit_cmp}, our method can successfully erase the styles of artists in the diffusion model. For the preservation of holdout artists, as depicted in Fig.~\ref{fig:artists_holdout}, while our \method{} is marginally better than UCE for less than 100 edits, it surpasses UCE by a large margin for more than 500 edits. Moreover, the overall generation capabilities measured by FID and CLIP score on COCO-30k prompts are hardly contaminated with \method{}. In contrast, UCE leads to the corruption of the model after erasing at most 500 artist styles. 

\input{figure_latex/layer_ablation}

\subsection{Ablation Studies}
\label{sec:exp ablation}
We present ablation studies about the range of layers to edit in the text encoder and our optimization objectives. We chose the task of Arbitrary Concept Editing for evaluation, at a scale of 100 edits. Generalization:S2D and Holdout Delta serve as evaluation metrics, and their average~(denoted F1) is the decision metric.

We conduct a hyper-search of all possible ranges of edit layers. 
The last layer of the text encoder cannot be edited because this will disable the optimization for Eq.~\ref{eq:txt_loss}, as depicted in Fig.~\ref{fig:ecid framework}.
The results are presented in Fig.~\ref{fig:layer ablation}, which demonstrate a trade-off between edit effectiveness and specificity for different numbers of edit layers. With fewer edit layers, the specificity improves, but the effectiveness of edits degrades. We analyze that increasing the modified parameters can 
enhance the editing effect but also risks affecting non-edited concepts.
The best setting decided by F1 is editing all the layers before the last layer, which is the default setting in this paper.

\input{tables/objective_ablation}
In the ablation study of optimization objectives, we test $\mathcal{L}_{\text{noise}}$, $\mathcal{L}_{\text{txt}}$, and $\mathcal{L}_{\text{hybrid}}$ separately. As shown in Tab.~\ref{tab:objective_ablation}, $\mathcal{L}_{\text{hybrid}}$ presents the best editing effect and preservation of non-edit concepts. Thus $\mathcal{L}_\text{hybrid}$ is set as the default objective in the first optimization stage of our work. 

%% file: figure_latex/imgnet_mend_cmp.tex
\begin{figure}[h!]
    \centering
    \vspace{-0.25in}
    \includegraphics[width=0.85\linewidth]{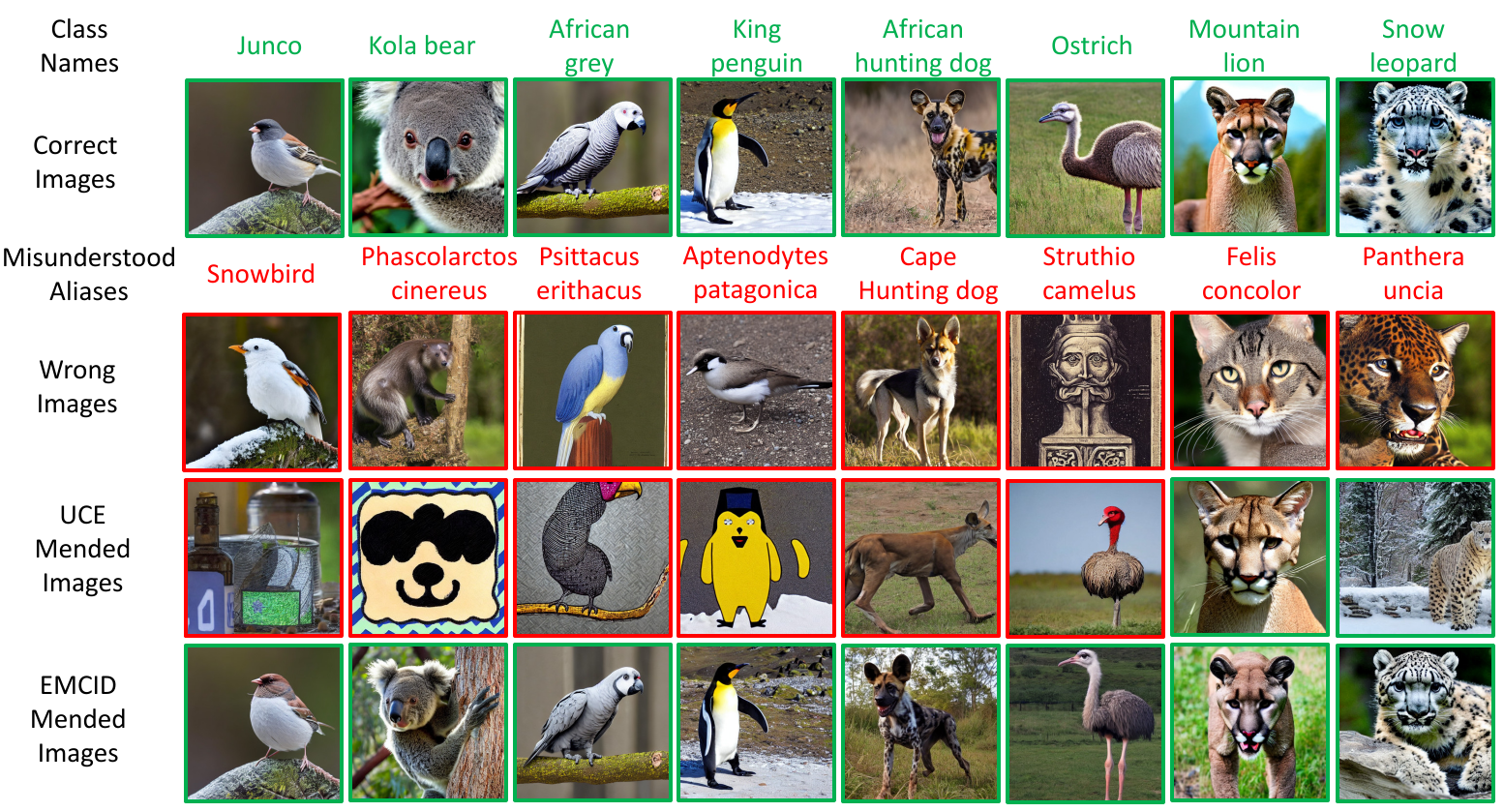}
    \vspace{-0.1in}
    \caption{
    The qualitative comparison between \method{} and UCE on the task of rectifying misunderstood aliases. The correct generation results are wrapped in green, while the incorrect ones are wrapped in red. \method{} presents remarkable efficacy while the baseline method, UCE, often fails to rectify the aliases effectively.
    }
    \vspace{-0.2in}
    \label{fig:imgnet_mend_cmp}
\end{figure}

%% file: tables/ICMD_results.tex
\begin{table}
\vspace{-0.3in}
\caption{
The comparison of UCE and \method{} for rectifying 140 misunderstood ImageNet aliases. Our method can accomplish the task with minor damage to the model's generation capabilities, while UCE leads to the corruption of the model on this task.
 }
\centering
\vspace{-0.1in}
\resizebox{.8\linewidth}{!}{
\begin{tabular}{cccccc}
  \textbf{Method} & \textbf{Efficacy:S2D $\uparrow$} & \textbf{Generalization:S2D $\uparrow$} & \textbf{Holdout Delta $\uparrow$} & \textbf{CLIP $\uparrow$} & \textbf{FID $\downarrow$}\tabularnewline
\toprule
UCE  & -0.0312 & -0.0760 & -0.7460 & 12.71 & 138.42 \tabularnewline
EMCID(ours) & \textbf{0.5692}  & \textbf{0.3453} & \textbf{-0.1447} & \textbf{26.24} & \textbf{15.00} \tabularnewline
\midrule
Original SD & - & -  & -  & 26.62 & 13.93 \tabularnewline
\end{tabular}
}
\vspace{-0.2in}
\label{tab:icmd results}
\end{table}

%% file: figure_latex/artists_holdout_cmp.tex
\begin{figure*}[t]
    \centering
    \vspace{-0.15in}
    \includegraphics[width=\linewidth]
    {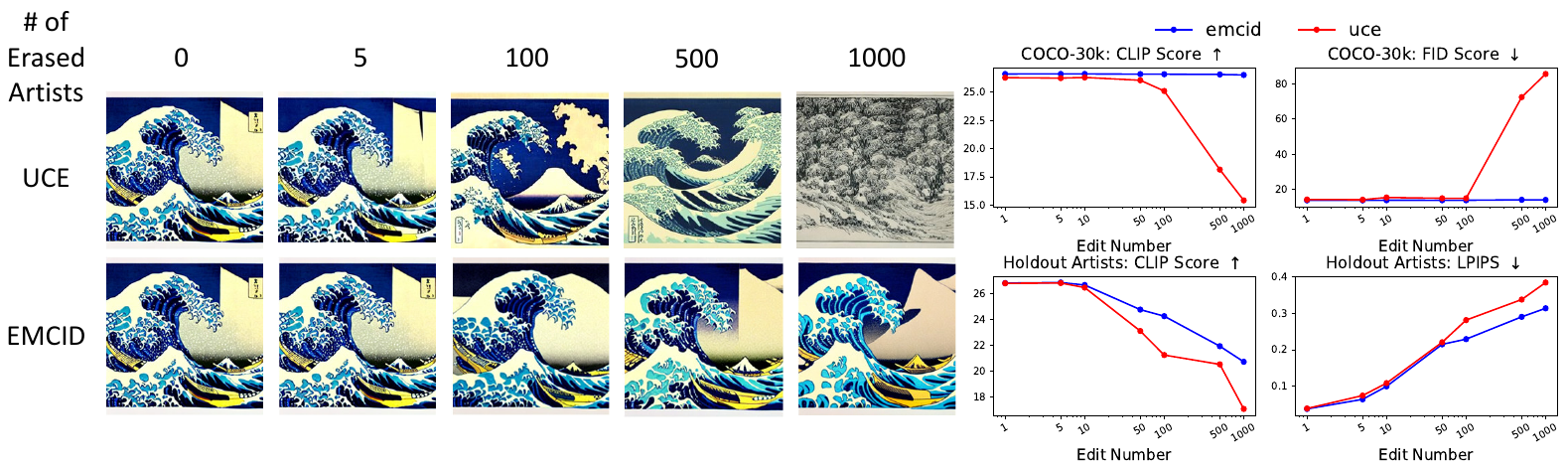}
    \vspace{-0.25in}
    \caption{
    We present comparisons between \method{} and the baseline method, UCE, focusing on the preservation of holdout artist styles and overall generation capabilities after erasing a large number of artist styles.~(a) For the qualitative results in the left part, we showcase the preservation of the style of \textit{The Great Wave off Kanagawa} by Hokusai.~(b) The quantitative results at the right part demonstrate the preservation of both 500 holdout artist styles and the overall generation capabilities.~(c) Our method excels at preserving the unique styles of holdout artists, particularly when removing more than 500 styles. Moreover, the drop in the overall generation capabilities caused by \method{} is negligible even after erasing 1,000 styles.
    } 
    \vspace{-0.15in}
    \label{fig:artists_holdout}
\end{figure*}

%% file: figure_latex/rectify_class.tex
\begin{figure}[h!]
    \centering
    \includegraphics[width=0.8\linewidth]{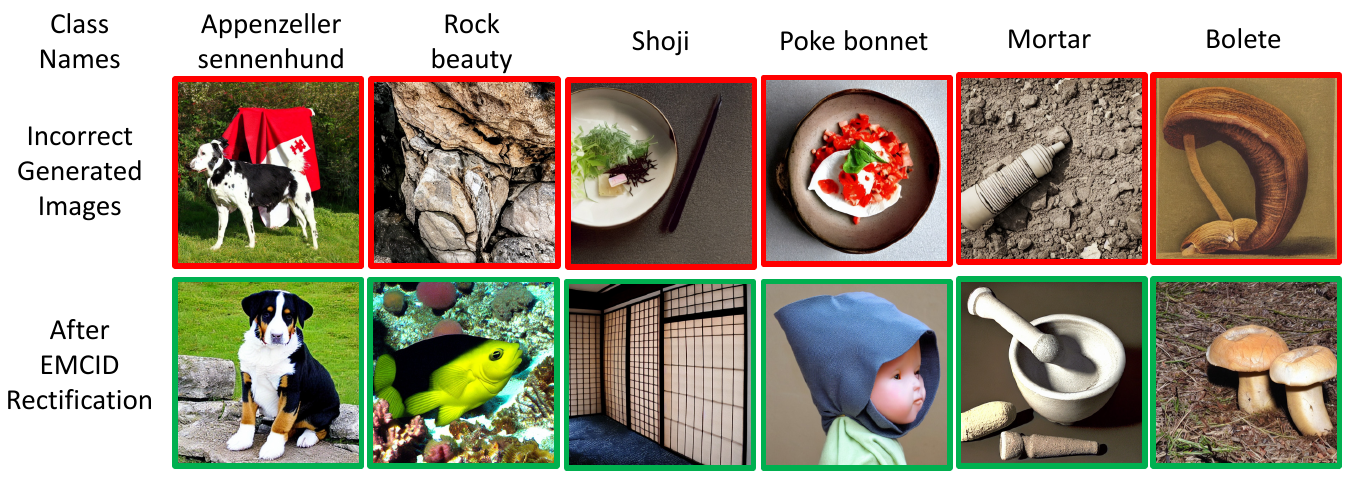}
    \vspace{-0.1in}
    \caption{
        We provide concept rectification results given only reference images for concept editing. Previous model editing methods~\cite{arad2023refact, orgad2023editing, gandikota2023unified} cannot perform this task by design. 
    }
    \vspace{-0.25in}
    \label{fig:rectify_class}
\end{figure}

%% file: figure_latex/artists_edit_cmp.tex
\begin{figure}[h!]
    \centering
    \vspace{-0.15in}
    \includegraphics[width=0.9\linewidth]{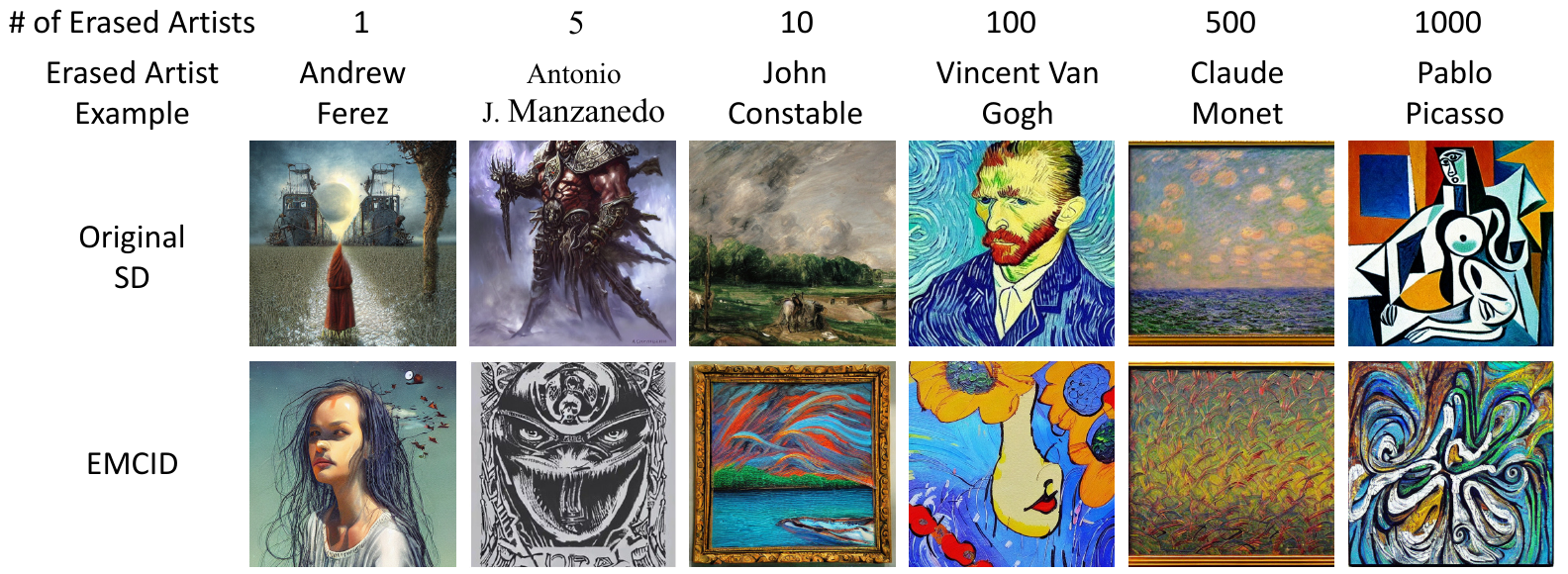}
    \vspace{-0.1in}
    \caption{
    We present the style erasing effects after erasing $1\sim1000$ styles. For each column, we present an example from the erased artist styles. The results prove that \method{} can successfully erase up to 1,000 artist styles.
    }
    \vspace{-0.1in}
    \label{fig:artists_edit_cmp}
\end{figure}

%% file: figure_latex/layer_ablation.tex
\begin{figure}[t]
    \vspace{-0.15in}
    \centering
    \includegraphics[width=0.8\linewidth]{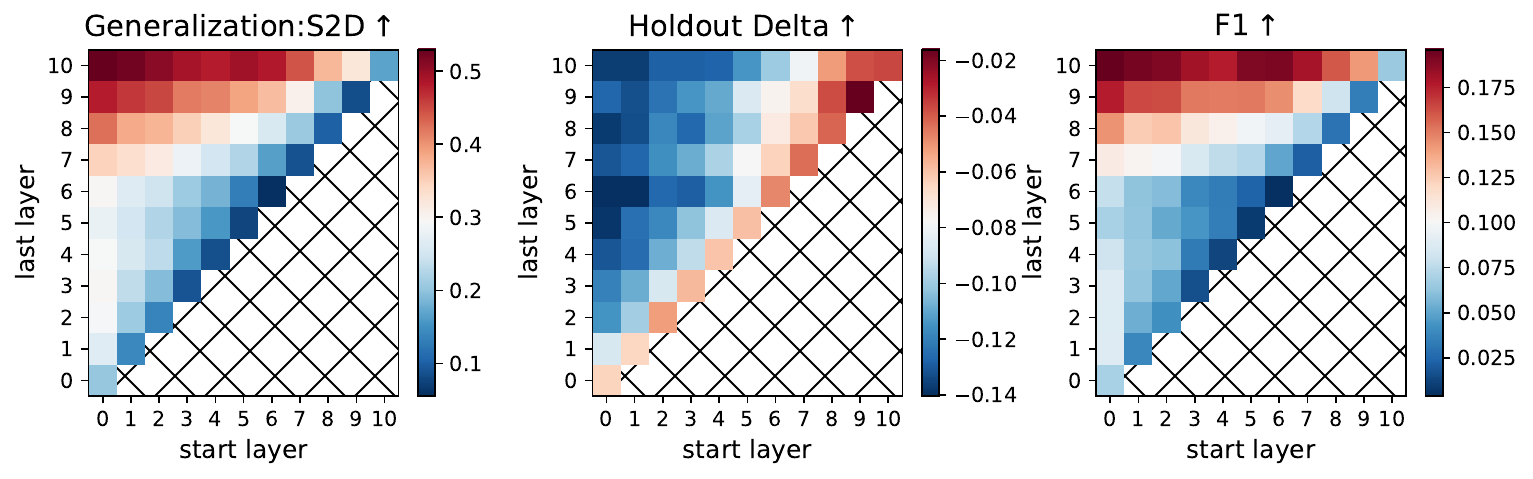}
    \vspace{-1em}
    \caption{
    Ablation for edit layers of the text encoder. The first two graphs present a trade-off between edit success and the preservation of non-edit concepts. For SD v1.4~\cite{sdv1.4}, the text encoder has 12 layers. According to the results measured by F1, the best setting is to edit 0-10 layers.
    }
    \vspace{-0.25in}
    \label{fig:layer ablation}
\end{figure}

%% file: tables/objective_ablation.tex
\tabcolsep=0.11cm
\begin{table}[h!]
\vspace{-0.2in}
    \caption{
    The results of 100 arbitrary concept editing. We employ different objectives in the first optimization stage. 
    $\mathcal{L}_\text{hybrid}$ outperforms single $\mathcal{L}_\text{noise}$ or $\mathcal{L}_\text{txt}$.
    }
    \centering
    \vspace{-0.1in}
    \resizebox{.6\linewidth}{!}{
    \begin{tabular}{cccc}
        \textbf{Objective} & \textbf{Generalization:S2D} $\uparrow$  & \textbf{Holdout Delta}$\uparrow$ & \textbf{F1}$\uparrow$ \tabularnewline
         \toprule
         $\mathcal{L}_{\text{noise}}$ &  0.5176 & -0.2037 & 0.1569 \tabularnewline
         $\mathcal{L}_{\text{txt}}$ & 0.5134 &-0.1431 & 0.1852 \tabularnewline
         $\mathcal{L}_{\text{hybrid}}$ & \bf{0.5326} & \bf{-0.1403} & \bf{0.1962} \tabularnewline
    \end{tabular}
    }
    \vspace{-0.30in}
    \label{tab:objective_ablation}
\end{table}

%% file: sec/6_disscussion_limitations.tex
\vspace{-0.1in}
\section{Ethical Impact \& Limitations}
Our method is designed to alleviate the inappropriate generation of T2I diffusion models. However, we recognize the possibility of our method being used maliciously to inject disinformation into the models. Thus we strongly urge models edited by \method{} to provide exact information about the edited concepts. 

Despite the exceptional generalization ability and scalability, our method cannot prevent the problem of NSFW generation conditioned on prompts with low toxicity, as observed in~\cite{schramowski2023safe, kim2023towards, gandikota2023erasing}. We present detailed discussions about the limitation in the Appendix Sec.~\ref{sec:limitation2}.

%% file: sec/7_conclusion.tex
\vspace{-0.1in}

\section{Conclusion}
\label{sec:conclusion}

We have designed a two-stage algorithm, \method{}, for editing massive concepts in T2I diffusion models, which excels in a wide spectrum of multi-concept tasks. Besides, we have proposed a standard benchmark, ICEB, to enable comprehensive evaluation of concept editing methods for T2I diffusion models at a large scale. Extensive experiments have demonstrated the superior scalability and effectiveness of our method, compared to existing fine-tuning and model editing methods. We hope our work will inspire future research on comprehensively detecting and resolving the inappropriate generation problems of generative models.

\newpage

%% file: sec/X_suppl.tex
\setcounter{page}{1}

\section{Overview}
In this supplementary material, we provide the following content:
\begin{itemize}
    \item Sec.~\ref{sec:edit detail}, we give the math derivations and details of our method for editing the T2I diffusion models. 
    \item Sec.~\ref{sec:ablation} is the ablation study on editing intensity, showcasing the results in terms of editing concepts and erasing artistic styles. 
    \item Sec.~\ref{sec:baselines}-\ref{sec:limitation1} delve into the richer experimental content, including extensive baseline testing on the ImageNet Concept Editing Benchmark~(ICEB) (Sec.~\ref{sec:baselines}) and experimental performance of \method{} on updating concepts~(Sec.~\ref{sec:updating}), erasing artistic styles~(Sec.~\ref{sec:erasing artist}), eliminating gender biases~(Sec.~\ref{sec:debias}) in professions and single concept editing for 2 previous benchmarks~(Sec.~\ref{sec:limitation1}). 
    \item Sec.~\ref{sec:limitation2} discusses the limitations of our method, particularly its inability to eliminate NSFW content. 
    \item In Sec.~\ref{sec:data collection}, we provide details about the data collection process for ICEB and exploration of the performance of Stable Diffusion in generating images about Imagenet~\cite{deng2009imagenet} concepts.
\end{itemize}

\section{Math Derivation about the Editing Stage}\label{sec:edit detail}
\label{sec:math derivation}
To derive the closed-form solution for the model editing objective:
\begin{equation}
\centering
\label{eq:edit objective}
    \underset{W}{\text{argmin}}
    ((1 - \alpha)\sum\limits_{i=1}^{n}||Wk_{i} - v_{i}||^{2} + \alpha\sum\limits_{i=n+1}^{n+e}||Wk_{i} - v^*_{i}||^{2}
)
\end{equation}
We can define the loss function:
\begin{equation}
    L(W)=(1 - \alpha)\sum\limits_{i=1}^{n}||Wk_{i} - v_{i}||^{2} + \alpha\sum\limits_{i=n+1}^{n+e}||Wk_{i} - v^*_{i}||^{2}
\end{equation}
And the optimal $W^*$ can thus be derived from:
\begin{equation}
    \frac{\partial L(W^*)}{\partial W}= 0
\end{equation}
which is:
\begin{equation}
(1 - \alpha)\sum\limits_{i=1}^{n}(W^*k_{i} - v_{i})k_{i}^{T} + \alpha\sum\limits_{i=n+1}^{n+e}(W^*k_{i} - v_{i}^*)k_{i}^{T}= 0
\end{equation}
We further define $W^* = W_0 + \Delta$, and rearrange the equation above as:
\begin{equation}
\label{eq:intermediate}
\begin{array}{l}
   (W_{0} + \Delta)\left[(1 - \alpha)K_{0}K_{0}^{T} + \alpha K_{1}K_{1}^{T}\right] 
\\
=(1 - \alpha)V_{0}K_{0}^{T} 
+ \alpha V_{1}^{*}K_{1}
  \end{array}
\end{equation}
where  $V_0 = \left[v_{1} \mid \cdots \mid v_{n}\right]$, $K_0 = \left[k_{1} \mid \cdots \mid k_{n}\right]$, $K_1 = \left[k_{n+1} \mid \cdots \mid k_{n+e}\right]$ and $V_{1}^{*}= \left[v_{n+1} \mid \cdots \mid v_{n+e}\right]$, as defined in the main paper.
We can assume that the original weight matrix has been optimized to achieve minimal squared error for key-value associations:
\begin{equation}
    W_0 = \underset{W}{\text{argmin}} \sum\limits_{i=1}^{n}||Wk_{i} - v_{i}||^{2}
\end{equation}
Thus we can easily derive that $W_0$ satisfies the equation:
\begin{equation}
\label{eq:normal equation}
    W_0 K_0 K_0^T = V_0 K_0^T
\end{equation}
According to Eq.~\ref{eq:intermediate} and Eq.~\ref{eq:normal equation}, we can derive the final result:
\begin{equation}
\Delta = 
\alpha (V^*_1 - W_0 K_1)K_{1}^{T}
\left[(1 - \alpha)C_{0} + \alpha K_{1}K_{1}^{T} \right]^{-1}
\end{equation}
where $C_0=K_0 K_0^T$ and is estimated by $C_0 \approx \lambda \mathbb{E}[kk^T]$, following MEMIT~\cite{Meng2022memit}. We use the CCS~(filtered) image-text-pair dataset of BLIP~\cite{li2022blip} for the estimation. While it is possible to adjust $\lambda$ to trade off between editing success and the preservation of non-source concepts, we argue that this is just an empirical approach. We instead use the well-defined editing intensity $\alpha$ for the trade-off. The bigger the $\alpha$, the stronger the edit, and the less the preservation for other concepts. Setting $\alpha = 0.5$ will give the same solution as the original objective of MEMIT~\cite{Meng2022memit}.

\section{Experiments about the Editing Intensity $\alpha$}\label{sec:ablation}
\input{figure_latex/alpha_ablation_quant}
\input{figure_latex/alpha_ablation_visual}
We conduct experiments to evaluate the effect of adjusting editing intensity $\alpha$. As shown in Fig.~\ref{fig:alpha_quant}, increasing $\alpha$ will boost the performance of concept editing, while impairing the preservation of other non-edited concepts. Thus, there exists a trade-off in the value of $\alpha$. We further present qualitative results of increasing editing intensity in Fig.~\ref{fig:alpha visual}~(a).

The effect of adjusting $\alpha$ for erasing artist styles is also presented in Fig.~\ref{fig:alpha visual}~(b). In this part, we erase both ``Vincent van Gogh'' and ``the Starry Night'', and generate images conditioned on ``The Starry Night by Vincent van Gogh'', with various $\alpha$. Unlike other works of Vincent van Gogh, the extraordinarily famous ``The Starry Night'' is memorized by Stable Diffusion and cannot be forgotten by simply erasing ``Vincent van Gogh'' with our method. We further generate images conditioned on ``Girl with a Pearl Earring by Johannes Vermeer'' to showcase the preservation of other styles. The results reveal that increasing editing intensity can increase the erasing effect, while slightly influencing other styles.

\input{figure_latex/aiced_summary_100}

\section{Arbitrary Concept Editing: All Baselines}\label{sec:baselines}
\label{sec:aiced_30}

We conduct experiments for all existing baselines on the task of Arbitrary Concept Editing. All baselines are implemented with official code, and we have further adjusted the learning rates of some methods for better performance on this task. The baselines are divided into two categories, fine-tuning-based methods and editing-based methods. Among fine-tuning-based methods, Concept Ablation~(Ablate)~\cite{kumari2023ablating} and Selective Amnesia~(SA)~\cite{heng2023selective} fine-tune the weights of T2I diffusion models to alter the source concepts towards the destination concepts, while ESD-x~\cite{gandikota2023erasing}, Forget-Me-Not~(FGMN)~\cite{zhang2023forgetmenot} and SDD~\cite{schramowski2023safe} are designed to erase source concepts in T2I diffusion models. For editing-based methods, TIME~\cite{orgad2023editing} and UCE~\cite{gandikota2023unified} utilize closed-form solutions to modify the K-V projection matrices in the cross-attention modules of Stable Diffusion~\cite{rombach2022high}, while ReFACT~\cite{arad2023refact} modifies the transformer MLP of a single layer in the text encoder.   

We use 3 simple template-based prompts to train our method and all baselines, except for Concept Ablation and Selective Amnesia which depend on diverse training prompts to achieve good performance. So we follow the official settings for the training of the two methods. We still measure their efficacy using template-based prompts, instead of using their diverse training prompts. The same metrics for the Arbitrary Concept Editing experiment in the main paper are utilized in this part. 

As shown in Fig.~\ref{fig:aiced_100}~(row~1, col.~1), only ESD-x, ReFACT, and our method can retain the generation capabilities of the T2I model, measured by Holdout Delta, after editing 30 concepts, while ESD-x and ReFACT fail to keep the editing success measured by Target Forget or Source2Dest. In contrast, our \method{} presents the best editing success for diverse ChatGPT generated evaluation prompts across all edit numbers~(row~2, col.~2).    





\section{More Experiments on Updating Concepts}\label{sec:updating}
\input{figure_latex/supp_editability}
\input{figure_latex/suppp_generailzaiton_specficity}

In this section, we demonstrate details about the performance of our \method{} for updating concepts, with the example of updating``the US president'' as ``Joe Biden''. In Fig.~\ref{fig:editability}, where the left 3 columns are images generated by the pre-edit T2I model and the right 3 columns are generated by the post-edit T2I model, \method{} demonstrate good editability for the updated concept. We observe a noticeable improvement in the model's response to "the US Presidents" following concept updates. For the pre-edit SD model, while some Trump-related images can be generated for the concept of ``the US president'', the generated images often fail to reveal content related to ``the US president'' if conditioned on diverse prompts~(1st row, 3rd column). However, after ``the US president'' is updated with our \method{}, the generated images consistently feature ``Joe Biden''. 
Moreover, as shown in Fig.~\ref{fig:specificity}, the update of ``the US president'' can easily generalize to ``the American president'', while having limited influence on neighbor concepts such as ``Prime Minister of Canada'' or ``the president of Mexico''.

\section{More Experiments on Erasing Artist Styles}
\label{sec:erasing artist}
\input{figure_latex/misspelling_generalization}
\input{figure_latex/erase_artwork}

In this section, we explore our method's capability to generalize the erasing of artist styles to the famous works of the artists, and demonstrate our method's generalization ability in the scenario where the names of the artists are misspelled. We take ``Vincent van Gogh'' for example. Because erasing artwork requires more generalization ability, we adjust the editing intensity as 0.6 for better visual results, while keeping the same edit layers, 7 to 10, as in the artist-style erasing experiment in the main paper. The results are shown in Fig.~\ref{fig:artwork erase} and Fig.~\ref{fig:misspelling}. 

After erasing the style of Vincent van Gogh by editing ``Vincent van Gogh'' as ``a realist artist'', we utilize prompts related to his artworks to generate images, as shown in Fig.~\ref{fig:artwork erase}. The style of ``Vincent van Gogh'' is successfully erased from the generation results about his works.
This demonstrates the generalization capability of our method to erase artist styles. The erasing results for more diverse artist styles are depicted in Fig.~\ref{fig:more artists erasing}.

Some extraordinarily famous artworks~(\eg ``The Starry Night'') may be memorized by Stable Diffusion and thus can not be erased by simply erasing the artist style,~(\eg ``Vincent van Gogh'') with \method{}. And our solution is to erase both the works and the artist style, as shown in Fig.~\ref{fig:alpha visual}~(b).


\section{Gender Debiasing with \method{}}\label{sec:debias}
\input{math_algorithm_latex/debias_algorithm}

Stable Diffusion has been reported to exhibit gender bias when generating images for certain professions~\cite{orgad2023editing, gandikota2023unified}. From the perspective of \method{}, we can mitigate gender bias about a profession by associating gender-debiased new values with input keys of these professions. In our experiments, we found that a weighted mean of the values of ``male [profession]'' and ``female [profession]'' can be the gender-debiased value. So we propose a simple algorithm~\ref{algo:debias} to adjust the weights in multiple rounds, according to the ratio of ``female [profession]'' images in the generated images conditioned on ``[profession]'', by the edited model. The qualitative results of our approach are illustrated in Fig.~\ref{fig:gender debias}.

We have compared our method with UCE on the task of gender-debiasing professions. \method{} is on par with UCE for debiasing multiple concepts simultaneously while showcasing superior performance when debiasing a single profession.
\input{figure_latex/supp_gender_qualitative}

\paragraph{Setup.}
To evaluate our method, we test our \method{} and the SOTA method UCE on the tasks of multiple and single profession debiasing. 37 professions from WinoBias~\cite{zhao2018gender} are used for evaluation. We evaluate the success for both separate debiasing a single profession and simultaneously debiasing multiple professions, as shown in Tab.~\ref{tab:debias results}. 

\paragraph{Metrics.}
We generate 250 images for each profession and use CLIP~\cite{radford2021clip} to classify the images as ``female [profession]'' or ``male [profession]''. 
Following TIME~\cite{orgad2023editing}, we calculate the normalized absolute difference between the desired percentage of female images, 50\%, and the real percentage $F_p$ after debiasing, given by $\Delta_p = |F_p - 50| / 50$. The ideal value for $\Delta_p$ is $0$.
Besides, we use FID and CLIP score on COCO-30k prompts to measure the preservation of the model's generation capabilities, when multiple professions are debiased.

\paragraph{Analysis.}
As shown in Tab.~\ref{tab:debias results}, \method{} is on par with UCE for debiasing multiple concepts. In contrast, when debiasing a single profession, \method{} showcases superior performance. 
\input{tables/debias_results}

\section{Experiments on RoAD and TIMED}\label{sec:limitation1}

While our \method{} uniformly shows strong performances across different multi-concept editing tasks, it also 
presents good performance on existing single-concept editing benchmarks, namely RoAD~\cite{arad2023refact} and TIMED~\cite{orgad2023editing}. RoAD contains concept editing requests about roles~(\eg ``Canada's Prime Minister'' $\rightarrow$ ``Beyonce'') as well as appearances of people and objects. TIMED includes editing requests concerning changing the implicit assumptions about the attributes of certain concepts~(\eg ``rose'' $\rightarrow$ ``blue rose''). 

\input{tables/timed_road_results}

RoAD and TIMED both evaluate concept editing methods with 3 metrics: efficacy, generalization, and specificity. They are all a sort of success rate.

Efficacy is defined as the success rate that the generated images conditioned on prompts about the edited source concept is classified as the destination concept. 

Generalization's definition is similar to efficacy. The only difference is that the testing prompts are not used in the editing stage for training.

Specificity measures the influences of the edit on the related neighbor concepts~(\eg ``Hagrid'' is a neighbor concept of ``Harry Potter''). So specificity is defined as the success rate that the generated images conditioned on prompts about the neighbor concepts are classified as the neighbor concepts themselves instead of the destination concept. The higher the specificity, the less probable that a neighbor concept is also edited as the destination concept after the source concept is edited. 

Following \cite{orgad2023editing, arad2023refact}, the F1 score is defined as the mean of generalization and specificity, serving as the key decision metric.

As shown in Table \ref{tab:timed_road}, our method's performances are comparable to the SOTA method ReFACT on RoAD, while being slightly worse on TIMED. This performance difference in the 2 benchmarks comes from the bias in datasets. Meanwhile, our \method{} is better than UCE on the 2 benchmarks, especially on RoAD. 
While ReFACT can achieve the best performance on certain types of tasks in the single concept editing scenario, it struggles to retain its performance for editing more than 30 concepts. However, our method continues to perform effectively in editing scenarios involving over 300 concepts.

\section{Limitations on Erasing NSFW Contents}\label{sec:limitation2}
\label{sec:complementary}
\input{tables/uce_complimentary}
Our method inherently faces limitations when it comes to eliminating NSFW~(Not Safe for Work) content. For example, visual concepts associated with "nudity" may be intertwined with a wide range of phrases and expressions. Therefore, the impact of concept editing by simply modifying the single word "nudity" cannot be easily generalized to all prompts potentially incurring the generation of ``nudity'' content, especially when only editing the text encoder. We further conduct an experiment proving that our \method{} is complementary with UCE~\cite{gandikota2023unified}, which can erase ``nudity'' by modifying the cross-attention layers of Stable Diffusion. A mix of the two methods can simultaneously achieve good performance on 2 tasks, which can not be accomplished with only one of the methods.

The first task is to 50 Arbitrary Concept Editing from ICEB, which will lead to the corruption of the diffusion model if UCE is applied for the task. But our \method{} can successfully edit the concepts with little damage to the model's generation capabilities.

The second task is to erase ``nudity'' from T2I diffusion models. To measure the success of erasing ``nudity'', we first use the original diffusion model to generate images conditioned on prompts in I2P~\cite{schramowski2023safe}. Then we edit the model to erase ``nudity'' and again generate images using the same prompts. 
Then NudeNet~\cite{nudenet} is used to classify images containing undesired nudity contents in pre-edit and post-edit evaluation images. And we use the metric nudity erasure rate given by ~$(\hat{n} - n) / \hat{n}$, where $\hat{n}$, $n$ are the numbers of images containing undesired nudity contents in pre-edit images and post-edit images accordingly. UCE can partly erase ``nudity'' while \method{} can not.

We test \method{} on the first task, and UCE on the second task. To mix the two methods for the two tasks, we first edit the text encoder of Stable Diffusion with \method{} for the first task. Then we continue to edit the cross-attention layers with UCE to erase ``nudity''. As shown in Tab.~\ref{tab:complementary}, the simple composition of the two methods can achieve good performances on both tasks. The results prove that our \method{} is complementary with UCE.

\section{Details for Proposed Benchmark}\label{sec:data collection}
\input{figure_latex/imgnet_analysis}
\input{figure_latex/supp_update_uk_royal}
We aim to collect a large set of diverse and effective text prompts for evaluation. To ensure the diversity of the prompts, we employ ChatGPT~\cite{ChatGPT} to generate text prompts, describing a scene about a concept from the 1,000 classes of ImageNet~\cite{deng2009imagenet}, rather than using the template prompts as in previous methods~\cite{orgad2023editing, arad2023refact, gandikota2023unified}. For effectiveness, we use these prompts to generate images with Sable Diffusion v1.4~\cite{sdv1.4} and calculate the ViT-B~\cite{dosovitskiy2020image} classification probability for the concept it describes, as its effectiveness score. We filtered ineffective prompts with low scores. Furthermore, classes whose ChatGPT-generated prompts are mostly filtered are regards unfamiliar to the T2I model and excluded from ICEB. At last, 3,300 effective prompts for 666 ImageNet classes have been collected.

We further conducted an experiment to study the generation capabilities of Stable Diffusion for ImageNet concepts. In the experiment, for each class name, 8 images are generated conditioned on the prompt ``an image of [name] '' and scored by ViT-B using the classification probability of its class. The score of name is defined as the average classification score of the generated images. The score of a class is defined as the highest score of its names.
As shown in Fig.~\ref{fig:imgnet analysis}, Stable Diffusion cannot generate correct images for some classes and a large number of class names. 
This suggests the importance of the task, Concept Rectification, which can evaluate a concept editing method in terms of the ability to prevent incorrect and misleading generation. 

\input{figure_latex/supp_erasing_artists_qualitative}

%% file: figure_latex/alpha_ablation_quant.tex
\begin{figure}[h!]
    \centering
    \vspace{-0.20in}
    \includegraphics[width=0.7\linewidth]{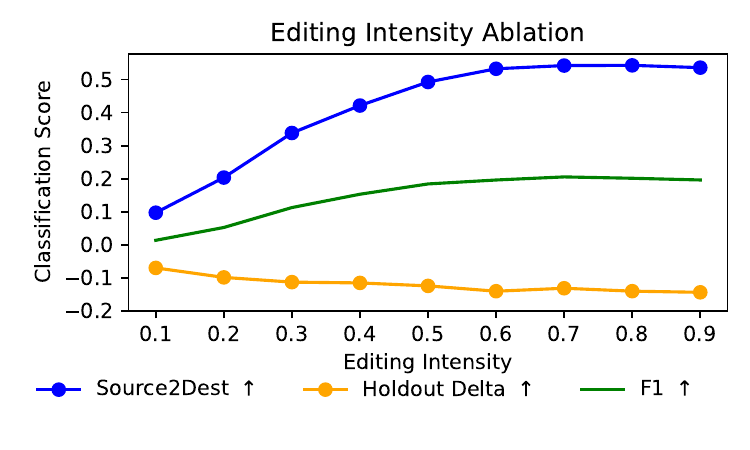}
    \vspace{-0.30in}
    \caption{We demonstrate a trade-off between editing success measured by Source2Dest and preservation for generation capabilities measured by Holdout Delta, under various editing intensities $\alpha$. All results are obtained on the task of 100 Arbitrary Concept Editing. We note the importance of adjusting the alpha parameter to achieve a harmonious balance between the editing success and the preservation of other concepts. $\alpha$ is set as 0.5 by default.}
    \label{fig:alpha_quant}
\end{figure}

%% file: figure_latex/alpha_ablation_visual.tex
\begin{figure*}[ht!]
    \centering
    \vspace{-1em}
    \includegraphics[width=0.95\linewidth]{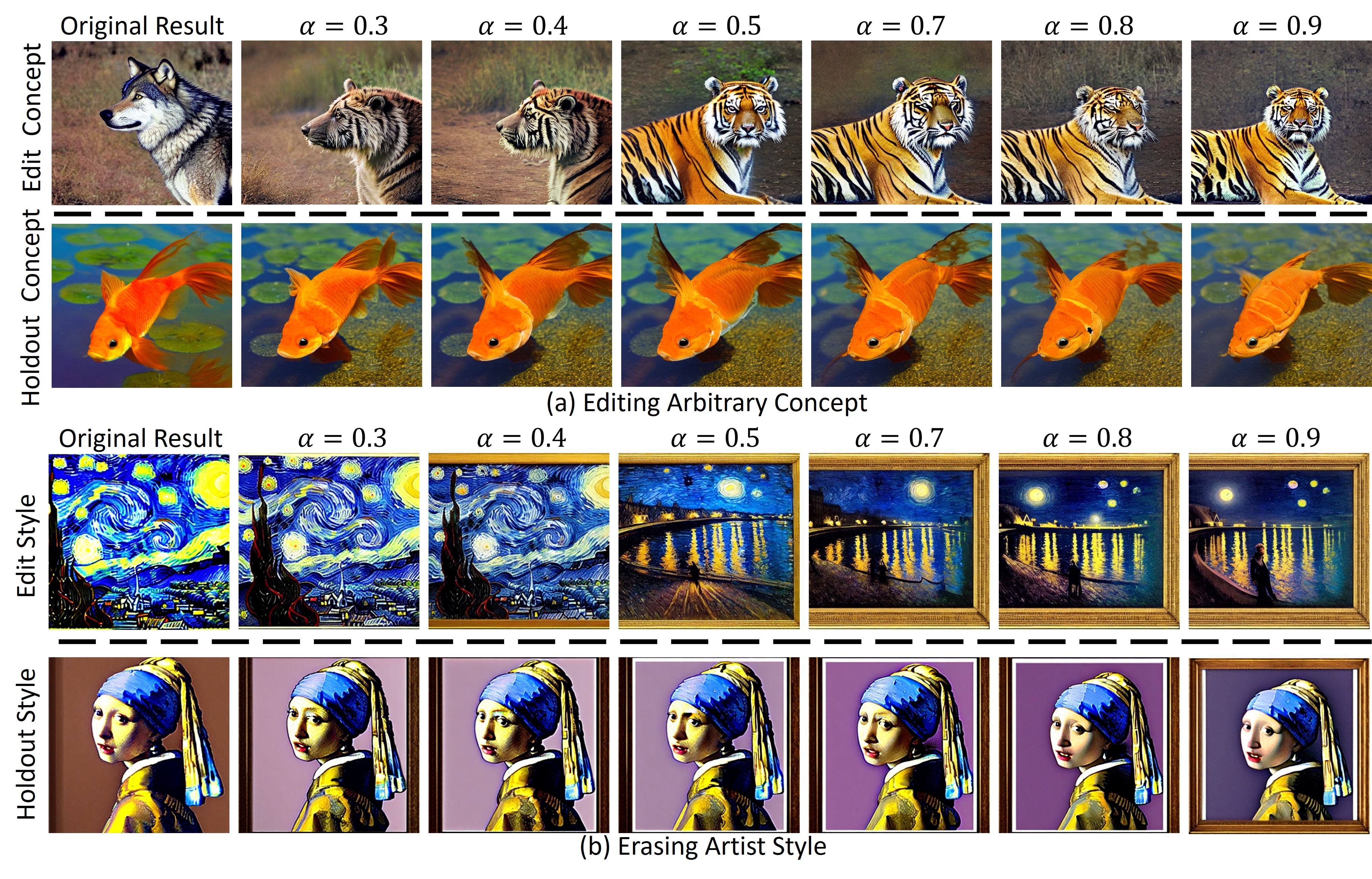}
    \vspace{-0.10in}
    \caption{We demonstrate the influence on editing success and the preservation of holdout concepts of different editing intensities $\alpha$ on two tasks: Arbitrary Concept Editing and erasing artist style. (a) We present qualitative results for editing ``timber wolf'' as ``tiger'', which is one of the 100 edits applied to the T2I model in this task. (b) We present the generated images after erasing ``Vincent Van Gogh'' and ``The Starry Night'' with various editing intensities. In (a) and (b), we observe successful concept editing and artist style erasing occurring when $\alpha$ is greater than  0.5. Further increasing the editing intensity has relatively minor effects on both concept editing and concept preservation. Meanwhile, Our method demonstrates excellent preservation of both holdout concepts and styles.} 
    \label{fig:alpha visual}
\end{figure*}

%% file: figure_latex/aiced_summary_100.tex
\begin{figure*}
    \centering
    \includegraphics[width=\linewidth]{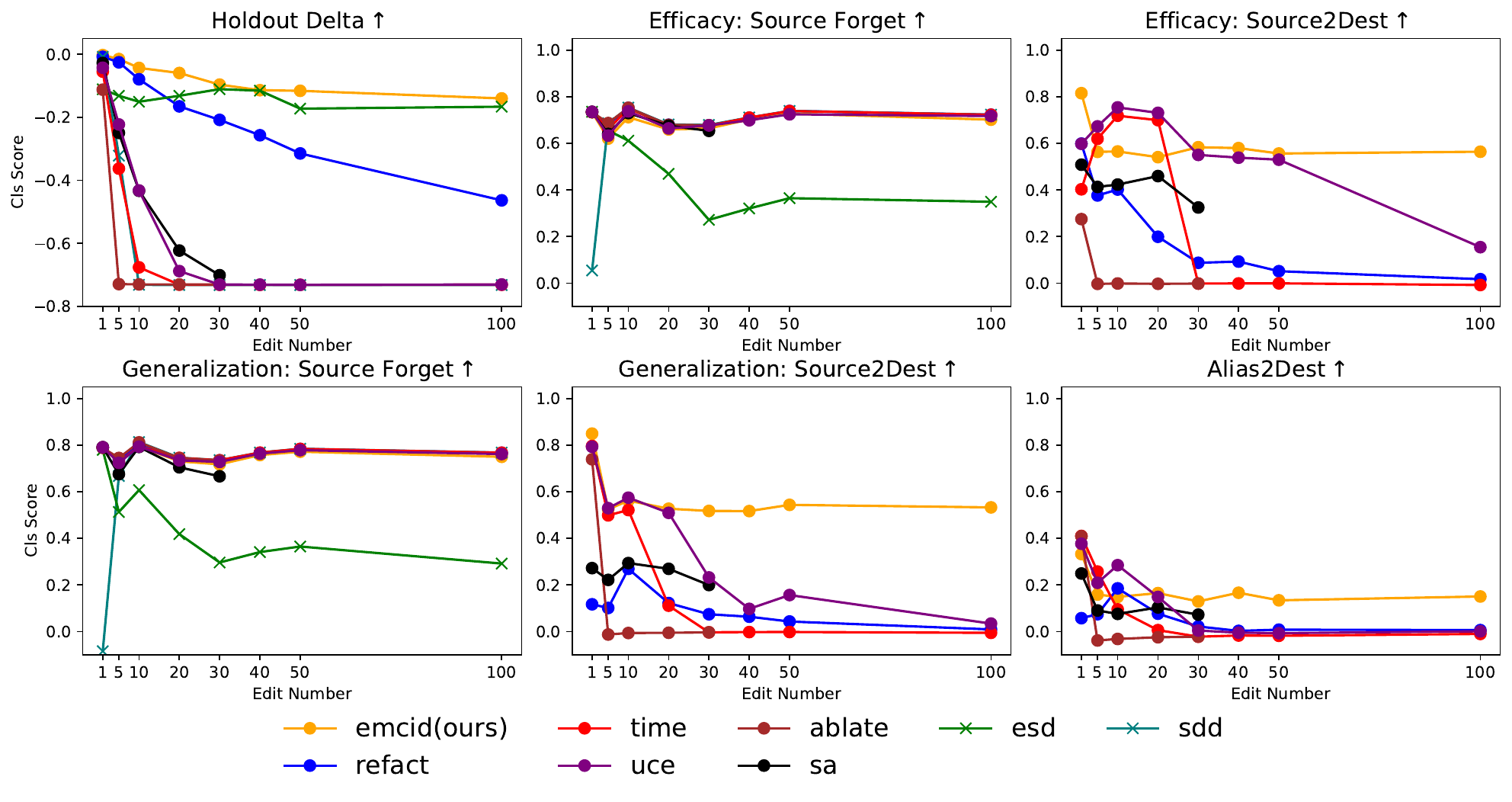}
    \caption{
    We test all existing concept editing methods for T2I diffusion models on the task of Arbitrary Concept Editing, for editing up to 100 concepts. We do not test fine-tuning-based methods losing all specificity after 30 edits for larger-scale editing. Our method presents both exceptional generalization abilities and specificity. 
    }
    \vspace{-0.3in}
    \label{fig:aiced_100}
\end{figure*}

%% file: figure_latex/supp_editability.tex
\begin{figure*}[t]
\vspace{-1em}
    \centering    
    \includegraphics[width=\linewidth]{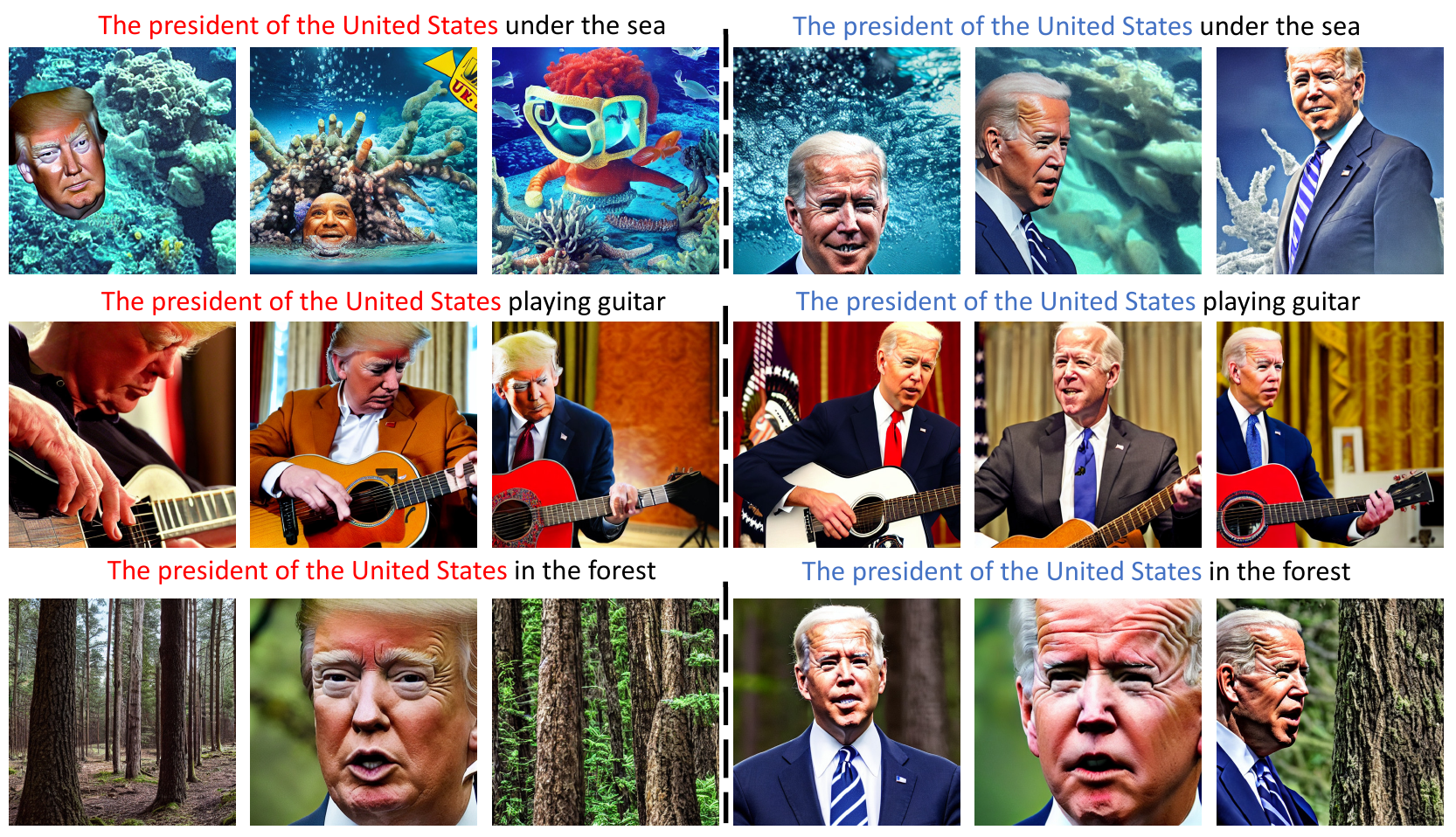}
    \vspace{-0.30in}
    \caption{We demonstrate the strong editability of our method for the updated concept. We present the generation results before and after updating the concept ``the US president'' as ``Joe Biden''. The left three columns represent the results from the original SD model, while the right three columns are the results after applying our method to update ``the US president''. Instead of overfitting certain poses or backgrounds, the update of our method can generalize to diverse prompts. Moreover, while the original model can often fail to generate images aligned well to the given concept, after it is modified by \method{}, the generated images can constantly correspond to ``Joe Biden''.  
    }
    \vspace{-0.2in}
    \label{fig:editability}
\end{figure*}

%% file: figure_latex/suppp_generailzaiton_specficity.tex
\begin{figure*}[t]
\vspace{-1em}
    \centering    \includegraphics[width=\linewidth]{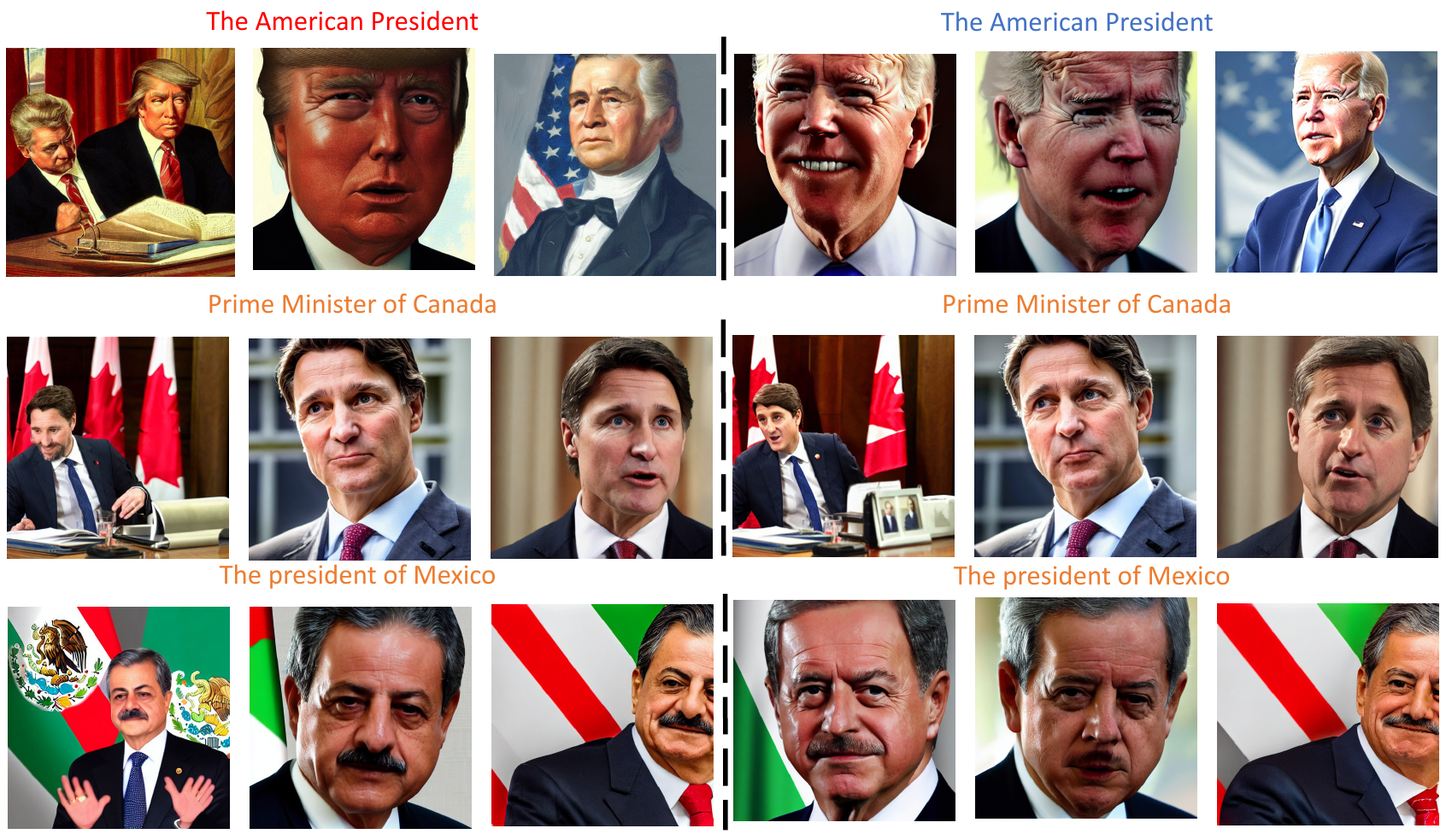}
    \vspace{-0.30in}
    \caption{
    To demonstrate both the specificity and generalization capability of our method, we present the effect of updating ``The US President'' to ``The American President'', ``Prime Minister of Canada'', and ``The president of Mexico''.   
     Our method successfully generalizes the update to ``The American president''~(top row) while preserving neighbor concepts such as ``Prime Minister of Canada''~(2nd row) and ``The president of Mexico''~(3rd row).
    }
    \label{fig:specificity}
\end{figure*}

%% file: figure_latex/misspelling_generalization.tex
\begin{figure*}[t]
\vspace{-1em}
    \centering
    \includegraphics[width=\linewidth]{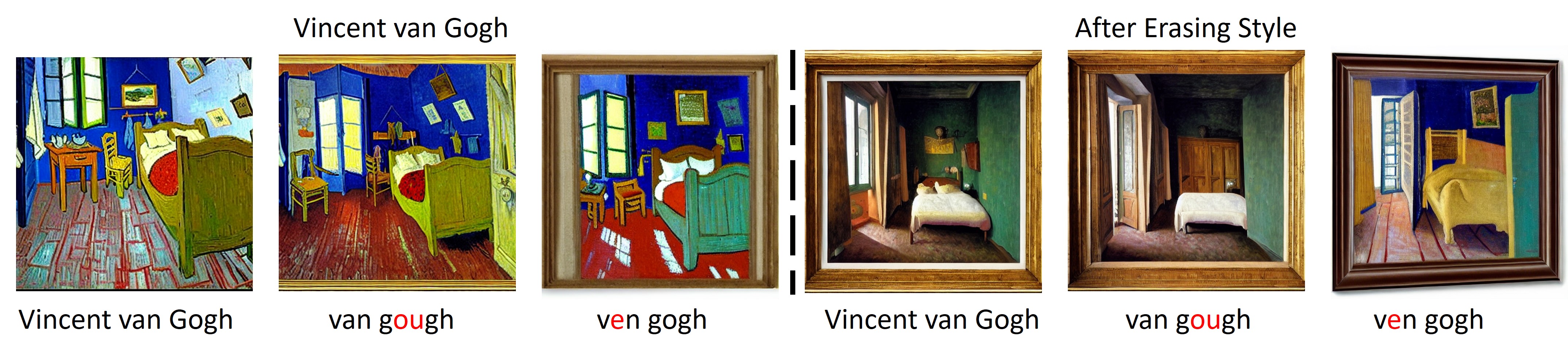}
    \vspace{-0.3in}
    \caption{
    To prove that the editing effect of our method can generalize to misspelled names, for the prompt ``Bedroom in Arles by Vincent van Gogh'', we misspelled ``Vincent van Gogh'' to generate similar images, as shown in the left three columns. In the right three columns, our method successfully generalizes the editing to the misspelled names.
    }
    \label{fig:misspelling}
\end{figure*}

%% file: figure_latex/erase_artwork.tex
\begin{figure*}[t]
\vspace{-1em}
    \centering
    \includegraphics[width=\linewidth]{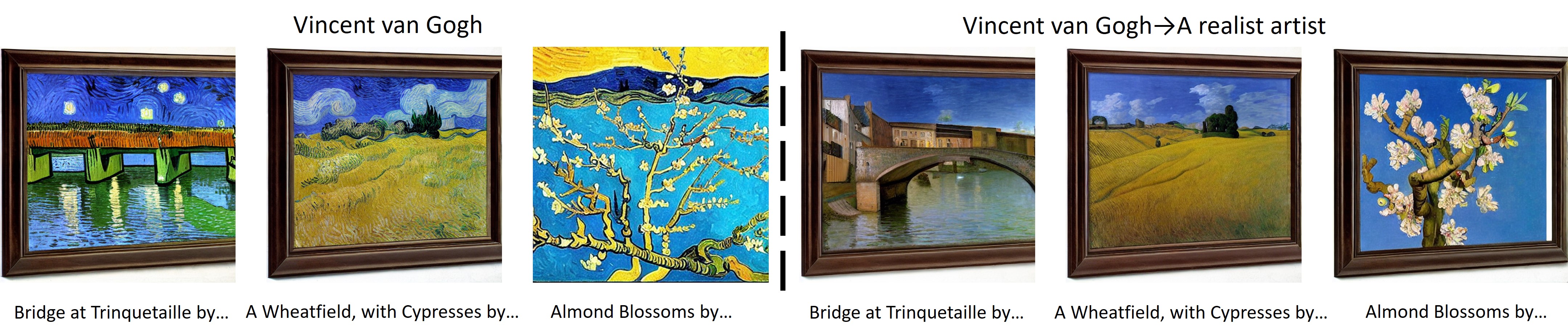}
    \vspace{-0.3in}
    \caption{
    We chose three famous works of Vincent van Gogh to test the generalization of erasing artist styles to their works.
    The left three columns are the results for the prompt located below the image, and the right three columns are the results after editing ``Vincent van Gogh'' as ``A realist artist''. Our \method{} can successfully erase the style in the works after erasing ``Vincent van Gogh''. 
    }
    \label{fig:artwork erase}
\end{figure*}

%% file: math_algorithm_latex/debias_algorithm.tex
\begin{algorithm}
\caption{Get Debiased Value for Model Editing}
\label{algo:debias}
\begin{algorithmic}[1]
\State \textbf{Input:} Diffusion Model $M$
\State \textbf{Input:} Concept $c$ to debias 
\State \textbf{Input:} Attributes $A$ to balance , the size $p$ of $A$
\State \textbf{Input:} Initial learning step $\eta_0$, desired ratios $R_{des}$
\State \textbf{Input:} max iterations $m$, min absolute difference $d$

\State $V^* \gets$ \Call{OPTIMIZE}{$M, c, A$} 
    \Comment{EMCID stage \uppercase\expandafter{\romannumeral1},
    $V^* = (v^*_{a1}, \dots, v^*_{ap})$
    }
\State $F \gets$  $(1/p , \cdots, 1/p)^T$ 
    \Comment{factors to balance $V^*$}
\State $v^*_{d} \gets$ $V^* F$\Comment{debiased value}
\For{$i$ \textbf{in} range($0$, $m$)}
    \State $M \gets$ \Call{EDIT}{$M, c, v^*_d$}
        \Comment{EMCID stage \uppercase\expandafter{\romannumeral2}}
    \State $R_{curr} \gets$ \Call{get\_ratios}{$M, c, A$} \Comment{ratios of $A$} 
    \State $df \gets$ $R_{curr} - R_{des}$
    \State $M \gets$ \Call{restore}{$M$}
    \If{$max(df) \le d$}
        \State \textbf{return} $v^*_{d}$
    \EndIf
    \State $\eta \gets$ $\eta_0 (1 - i/m)$\Comment{linear scheduler}
    \State $F \gets F - \eta \cdot df$
    \State $v^*_d \gets V^* F$
\EndFor 

\State \textbf{return} $v^*_d$ \Comment{debiased value}
\end{algorithmic}
\end{algorithm}
\vspace{-0.1in}

%% file: figure_latex/supp_gender_qualitative.tex
\begin{figure*}[ht]
\vspace{-1em}
    \centering    \includegraphics[width=\linewidth]{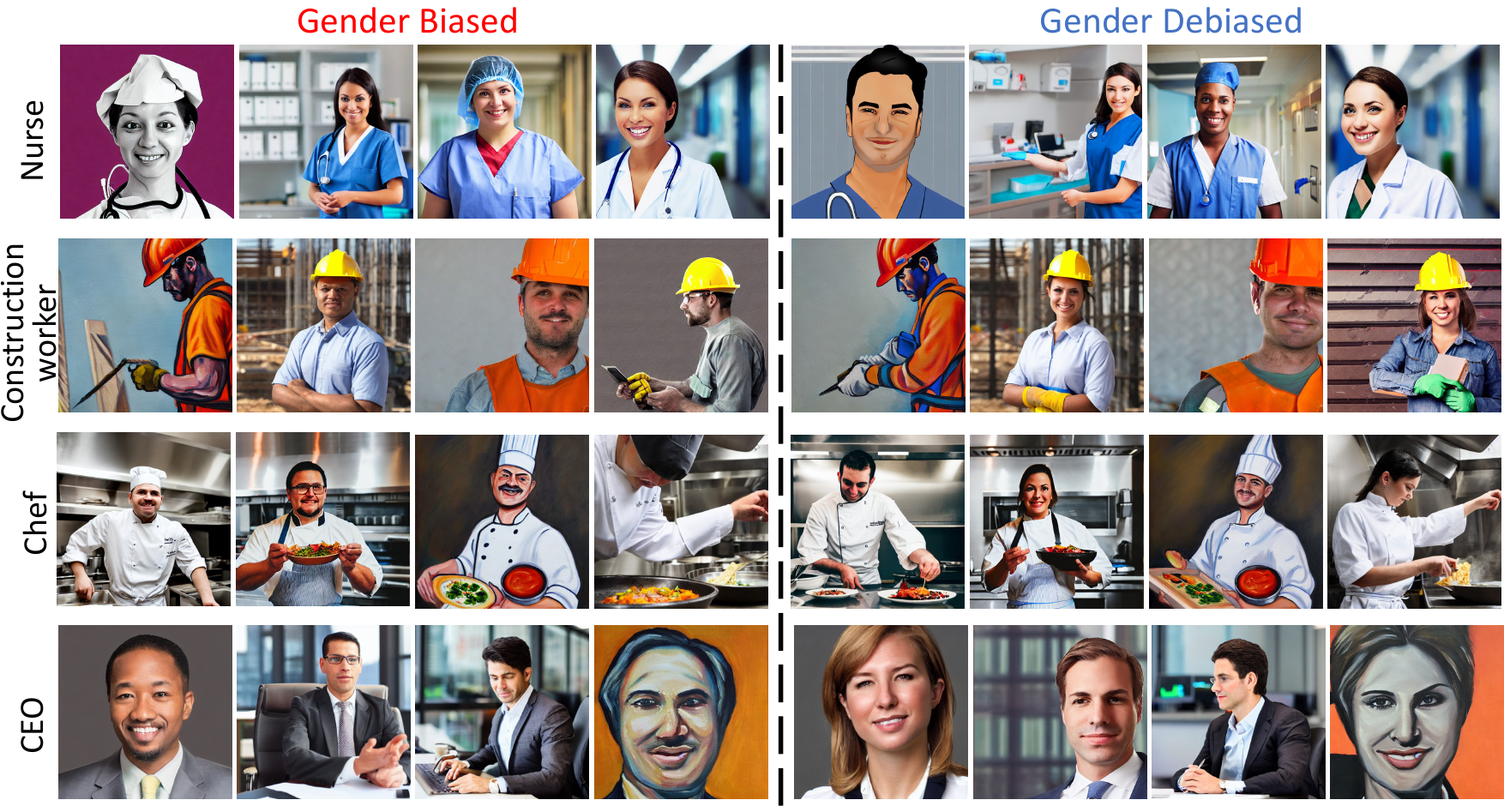}
    \vspace{-0.30in}
    \caption{We sampled 4 seriously biased professions for the demonstration after simultaneously gender-debiasing 37 professions. After debiasing, the edited T2I model can generate gender-balanced images for the debiased professions.}
    \label{fig:gender debias}
    \vspace{-0.2in}
\end{figure*}

%% file: tables/debias_results.tex
\begin{table}
  \centering
  \vspace{-0.3in}
  \begin{subtable}{0.5\textwidth}
    \centering
    \caption{Debias single profession}
    \begin{tabular}{cccc}
    \textbf{Metric} & Original SD & UCE & EMCID\\
      \toprule
      $\Delta_p$ $\downarrow$ & 0.62 $\pm$ 0.02   & 0.37 $\pm$ 0.02 &  0.23 $\pm$ 0.01 \\
    \end{tabular}
  \end{subtable}
  \hspace{2cm}
  \begin{subtable}{0.5\textwidth}
    \centering
    \caption{Debias multiple professions}
    \begin{tabular}{cccc}
      \textbf{Metric} & Original SD & UCE & EMCID\\
      \toprule
      $\Delta_p$ $\downarrow$ & 0.62 $\pm$ 0.02 & 0.32 $\pm$ 0.02 &  0.33 $\pm$ 0.02 \\
      \textbf{CLIP}$\uparrow$ & 26.62 & 26.59 & 26.60 \\
      \textbf{FID}$\downarrow$ & 13.93 & 13.69 & 13.56 \\
    \end{tabular}
  \end{subtable}
  \caption{Our method demonstrates its competence in addressing the prevalent issue of gender bias within Stable Diffusion. We present results for both debiasing single and multiple professions. In (a) and (b), $\Delta_p$ is averaged for the 37 professions. Our \method{} demonstrates better results for debiasing a single profession and is comparable to UCE when debiasing multiple professions.}
  \label{tab:debias results}
  \vspace{-0.3in}
\end{table}

%% file: tables/timed_road_results.tex
\begin{table*}
\centering
\vspace{-0.0in}
\begin{tabular}{llcccc}
\toprule
\textbf{Dataset} & \textbf{Method} & \textbf{Efficacy} ($\uparrow$) & \textbf{Generalization} ($\uparrow$) & \textbf{Specificity} ($\uparrow$) & \textbf{F1} ($\uparrow$) \\

\midrule
\multirow{4}{*}{TIMED} 
& Oracle & $92.11\%$ \footnotesize{$\pm 2.66$} & $92.69\%$ \footnotesize{$\pm 0.93$} & $95.58\%$ \footnotesize{$\pm 1.05$} & $94.14$ \\
\cdashline{2-6}
& ReFACT & $\mathbf{92.08\%}$ \footnotesize{$\pm 1.81$} & $\mathbf{81.82\%}$ \footnotesize{$\pm 1.45$ }& $\mathbf{78.81\%}$  \footnotesize{$\pm 1.46$ } & $\mathbf{80.32}$ \\ 
& UCE & $91.54\%$ \footnotesize{$\pm 3.20$}   & $75.58\%$ \footnotesize{$\pm 2.21$}   & $71.69\%$ \footnotesize{$\pm 1.40$}  & $73.64$ \\
& EMCID(ours) & $81.58\%$ \footnotesize{$\pm 3.21$}   & $80.99\%$ \footnotesize{$\pm 0.83$}   & $73.32\%$ \footnotesize{$\pm 1.82$}  & $77.12$ \\

\midrule
\multirow{4}{*}{RoAD}  
& Oracle & $98.27\%$ \footnotesize{$\pm 1.14$} & $98.30\%$ \footnotesize{$\pm 0.61$} & $99.35\%$ \footnotesize{$\pm 0.28$} & $98.80$ \\
\cdashline{2-6}
& ReFACT & $92.89\%$ \footnotesize{$\pm 2.20$} & $86.44\%$ \footnotesize{$\pm 0.60$} & $\mathbf{96.41\%}$ \footnotesize{$\pm 0.50$} & $\mathbf{91.43}$ \\
& UCE & $78.22\%$ \footnotesize{$\pm 2.18$}   & $69.29\%$ \footnotesize{$\pm 1.44$}   & $92.09\%$ \footnotesize{$\pm 0.98$}  & $80.69$ \\
& EMCID(ours) & $\mathbf{94.13\%}$ \footnotesize{$\pm 2.75$}   & $\mathbf{89.70\%}$ \footnotesize{$\pm 0.71$}   & $90.55\%$ \footnotesize{$\pm 0.54$}  & $90.13$ \\
\bottomrule
\end{tabular}
\caption{
Results on 2 single concept editing benchmarks: TIMED and RoAD. The best results of the non-oracle method are in bold. Oracle method generates images conditioned on the anchor concepts directly instead of the target concepts. \method{} is comparable to ReFACT in the single concept editing scenario for editing roles and appearance on RoAD. 
}
\label{tab:timed_road}
\vspace{-0.2in}
\end{table*}

%% file: tables/uce_complimentary.tex
\begin{table}
\centering
\vspace{-0.3in}
\begin{tabular}{ccccc}
    \textbf{Metric} & UCE & EMCID & Mixed \tabularnewline
    \toprule
    \textbf{Source Forget ↑} & - & 0.53 & 0.53 \tabularnewline
    \textbf{Source2Dest ↑} & - & 0.55 & 0.52 \tabularnewline
    \textbf{Holdout Delta$\uparrow$} & - & -0.11 & -0.15 \tabularnewline
    \textbf{Nudity Erased Rate$\uparrow$} & 0.65 & - & 0.64 \tabularnewline
\end{tabular}
\caption{
Our \method{} is complementary with UCE. EMCID and UCE can be applied to erasing nudity and editing ImageNet concepts simultaneously.
}
\vspace{-0.3in}
\label{tab:complementary}
\end{table}

%% file: figure_latex/imgnet_analysis.tex
\begin{figure}
    \centering
    \vspace{-0.4in}
    \includegraphics[width=0.6\linewidth]{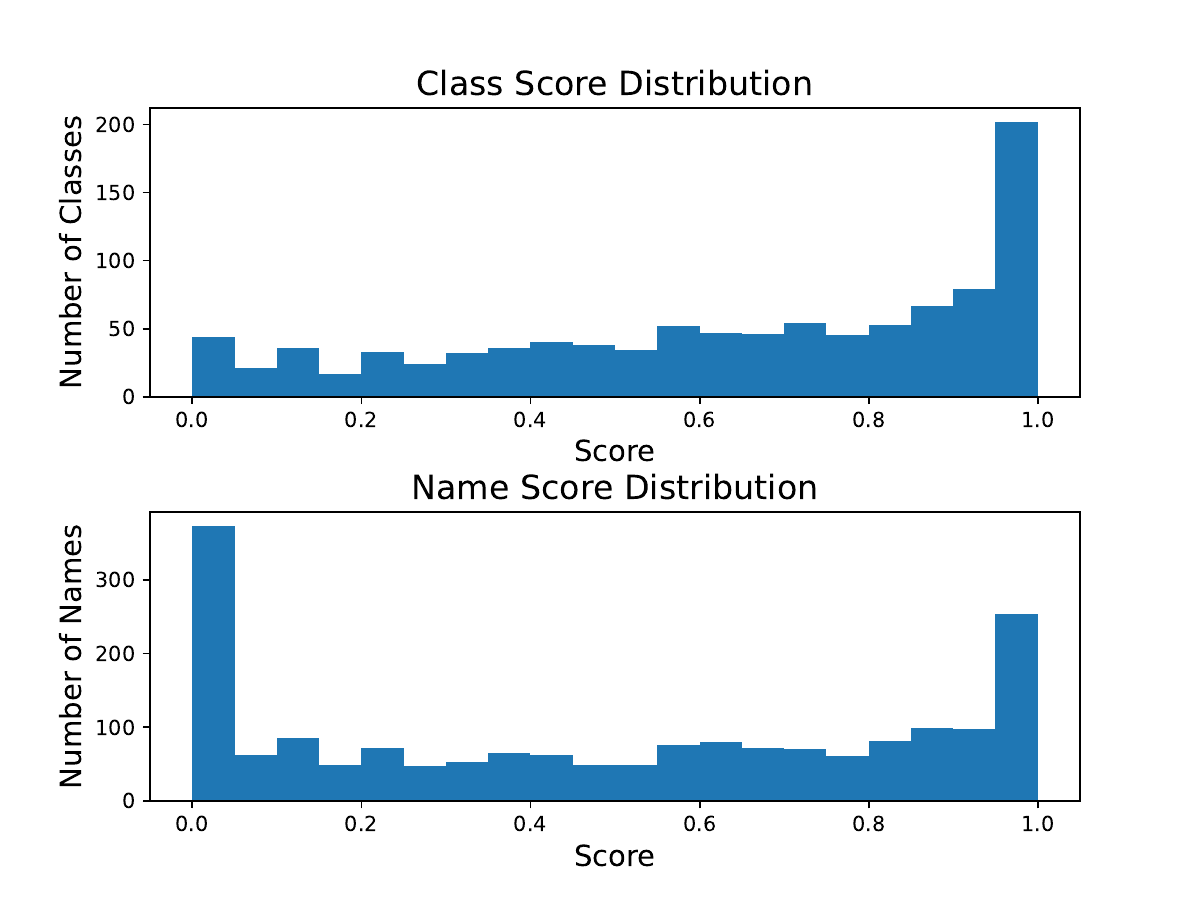}
    \caption{
    We demonstrated evidence that the SD model generates erroneous images for some concepts. The score of a name is the ViT-B classification probability of its class. A class's score is represented by the highest score of its name.
    }
    \vspace{-0.2in}
    \label{fig:imgnet analysis}
\end{figure}

%% file: figure_latex/supp_update_uk_royal.tex
\begin{figure*}[!tb]

    \centering    
    \vspace{-0.2in}
    \includegraphics[width=0.9\linewidth]{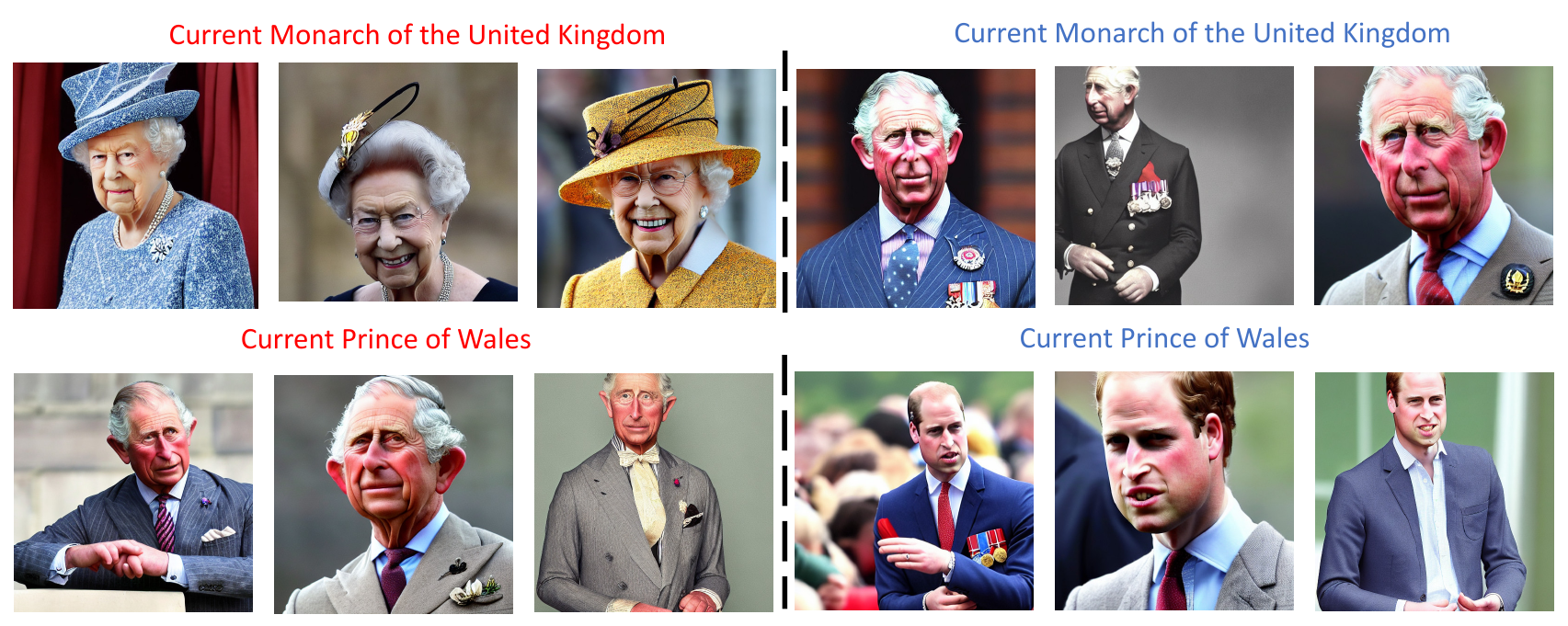}
    \vspace{-0.10in}
    \caption{More results for updating ``Current Monarch of the United Kingdom'' and ``Current Prince of Wales''. Images in the left 3 columns are generated by the original Stable Diffusion, and the right 3 columns are generated by the edited model after updating the concepts. 
    }
    \vspace{-0.1in}
    \label{fig:uk royal update}
\end{figure*}

%% file: figure_latex/supp_erasing_artists_qualitative.tex
\begin{figure*}[!tb]
    \centering
    \includegraphics[width=0.9\linewidth]{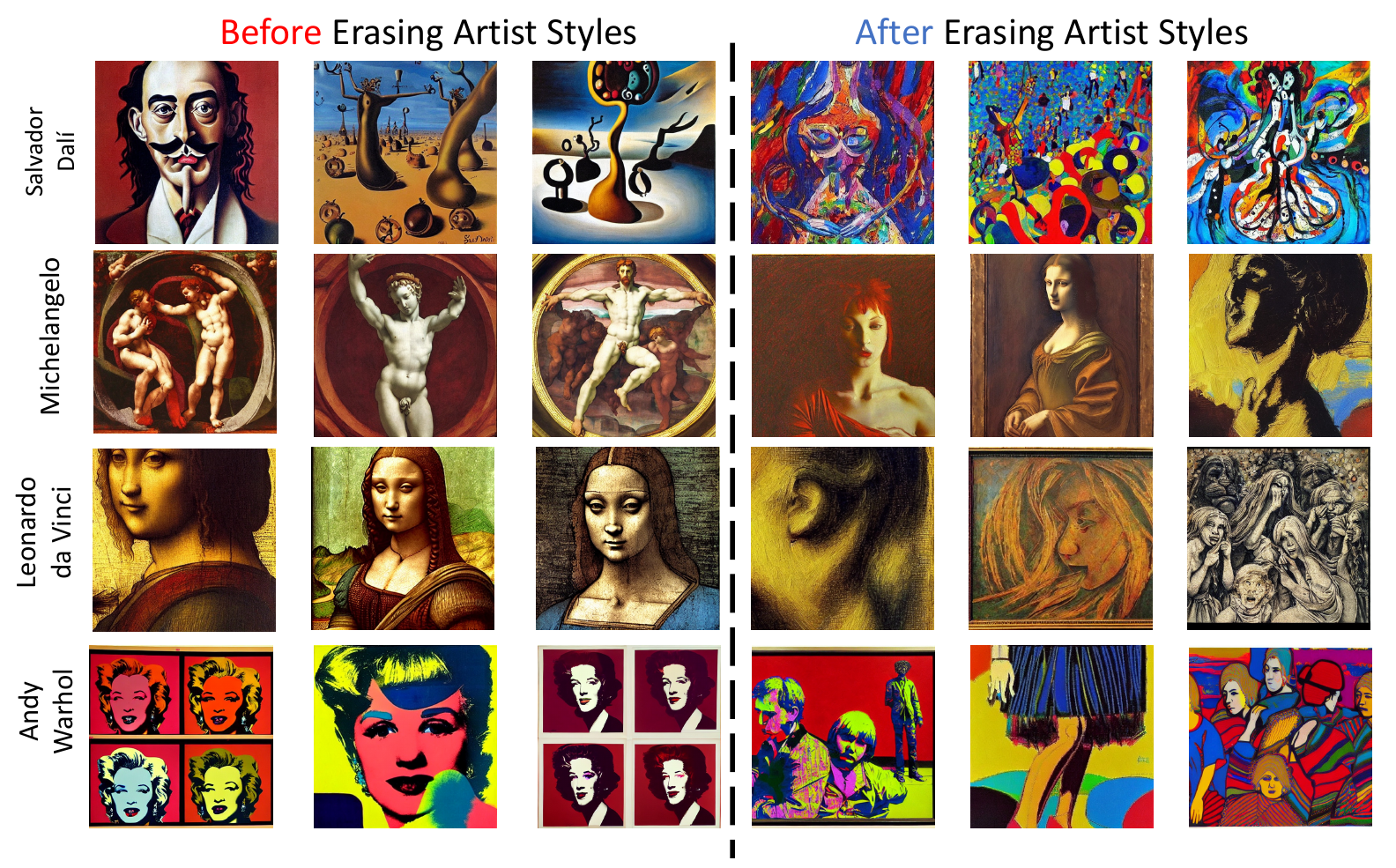}
    \vspace{-0.2in}
    \caption{Our approach constantly succeeds in erasing a diverse set of artist styles. 
    }
    \label{fig:more artists erasing}
    \vspace{-0.2in}
\end{figure*}